\documentclass{article} % For LaTeX2e
\usepackage{iclr2016_conference,times}
\usepackage[pdftex]{graphicx}
\DeclareGraphicsExtensions{.pdf,.png}
\usepackage{url}
\usepackage{ragged2e}
\usepackage{times}
\usepackage{array}
\usepackage{amsmath}
\usepackage{amssymb}
\usepackage{caption}
\usepackage{subcaption}
\usepackage[inline]{enumitem}

\newcommand{\source}{{\boldsymbol{\mathbf{s}}}}
\newcommand{\guide}{\boldsymbol{\mathbf{g}}}
\newcommand{\adv}{\boldsymbol{\mathbf{\alpha}}}
\newcommand{\neigh}[2]{\ensuremath{n_{#1} ({#2})}}
\newcommand{\Neighs}[2]{\ensuremath{\mathcal{N}}_{{#1}} ({#2})}

\newcommand{\class}[1]{C ({#1})}

\newcommand{\rank}[2]{\ensuremath{r_{#1} ({#2})}}
\newcommand{\rankdiff}[1]{\ensuremath{\Delta{r_{#1}}}}
\newcommand{\dlike}[2]{\ensuremath{\Delta L({#1}, {#2})}} %Difference in likelihood
\newcommand{\T}{\ensuremath{\top}} 
\definecolor{orange}{rgb}{1,0.5,0}
\newcommand{\yanshuai}[1]{\textcolor{red}{\textbf{[Y: #1]}}}
\newcommand{\david}[1]{\textcolor{blue}{\textbf{[D: #1]}}}

\newcommand{\comment}[1]{}

\newcommand{\beginsupplement}{%
\renewcommand{\thesection}{}% Remove section references...
\renewcommand{\thesubsection}{S\arabic{subsection}}%... from subsections
        \setcounter{table}{0}
        \renewcommand{\thetable}{S\arabic{table}}%
        \setcounter{figure}{0}
        \renewcommand{\thefigure}{S\arabic{figure}}%
     }
\usepackage[pagebackref=true,breaklinks=true,letterpaper=true,colorlinks,bookmarks=false]{hyperref}

\usepackage{onimage}
\tikzset{
     <new style name>/.style={
        fill=black,
        text=white,
        font=\fontfamily{phv}\selectfont\Small\bfseries
    }
}
\tikzset{
    image label/.style={
        every node/.style={
            fill=black,
            text=white,
            font=\fontfamily{phv}\selectfont\tiny\bfseries,
            anchor=south east,
            xshift=0.5cm,
            yshift=-0.5cm,
            at={(0,1)}
        }
    }
}
\title{Adversarial Manipulation of \\Deep Representations}
\author{
    Sara Sabour $^{* 1}$, Yanshuai Cao\thanks{The first two authors contributed 
    equally.}~~$^{1,2}$, Fartash Faghri$^{1,2}$ \& David J.  Fleet$^1$\\
    $^1$ Department of Computer Science,
    University of Toronto, Canada\\
    $^2$ Architech Labs, Toronto, Canada\\
    \texttt{\{saaraa,caoy,faghri,fleet\}@cs.toronto.edu}    }

\iclrfinalcopy % Uncomment for camera-ready version

\begin{document}

\maketitle

\begin{abstract}
\vspace*{-0.1cm}
We show that the image representations in a deep neural network 
(DNN) can be manipulated to mimic those of other natural images, 
with only minor, imperceptible perturbations to the original image. 
Previous methods for generating adversarial images focused on image 
perturbations designed to produce erroneous class labels.  Here we
instead concentrate on the internal layers of DNN representations, 
to produce a new class of adversarial images that differs qualitatively 
from others. While the adversary is perceptually similar to one image, 
its internal representation appears remarkably similar to a different 
image, from a different class and bearing little if any apparent 
similarity to the input.  Further, they appear generic and consistent 
with the space of natural images.  This phenomenon demonstrates the 
possibility to trick a DNN to confound almost any image with any other 
chosen image, and raises questions about DNN representations, as well 
as the properties of natural images themselves.

\vspace*{-0.1cm}
\end{abstract}

\section{Introduction}

\vspace*{-0.1cm}

%% \vspace*{-0.2cm}
%% \begin{figure}[h]
%% \centering
%% \includegraphics[width=.15\linewidth]{./imgs/technical-illustration-v3.png}
%% \caption{Summary} \label{fig:illustrate}
%% \vspace*{-0.2cm}
%% \end{figure}
%% \vspace*{-0.2cm}

Recent papers have shown that deep neural networks (DNNs) for image 
classification can be fooled, often using relatively simple methods to
generate so-called {\em adversarial images}
\citep{FawziEtalICLR2015, GoodfellowEtalICLR2015, GuRigazioNIPSWorkshop2014, 
NguyenEtAlCVPR2015, SzegedyElatICLR2014, Tabacof2015exploring}. 
% One such category of {\em adversarial images} is designed to disrupt image 
% classification, even though such adversaries differ almost imperceptibly 
% from the original source images 
The existence of adversarial images is important, not 
just because they reveal weaknesses in learned representations 
and classifiers, but because 1) they provide opportunities to explore 
fundamental questions about the nature of DNNs, e.g., whether they are 
inherent in the network structure per se or in the learned models, and 
2) such adversarial images might be harnessed to improve learning 
algorithms that yield better generalization and robustness 
\citep{GoodfellowEtalICLR2015, GuRigazioNIPSWorkshop2014}.

Research on adversarial images to date has focused mainly on disrupting 
classification, i.e., on algorithms that produce images classified with 
labels that are patently inconsistent with human perception. 
Given the large, potentially unbounded regions of feature space associated 
with a given class label, it may not be surprising that it is easy to
disrupt classification.
In this paper, in constrast to such {\em label adversaries}, we consider
a new, somewhat more incidious class of adversarial images, called
{\em feature adversaries}, which are confused with other images not 
just in the class label, but in their internal representations as well. 

Given a source image, a target (guide) image, and a trained DNN, we 
find small perturbations to the source image that produce an internal 
representation that is remarkably similar to that of the guide image, 
and hence far from that of the source.  
With this new class of adversarial phenomena  we demonstrate that it 
is possible to fool a DNN to confound almost any image with any other 
chosen image.
We further show that the deep representations of such adversarial images 
are not outliers per se.  Rather, they appear generic, indistinguishable 
from representations of natural images at multiple layers of a DNN.
This phenomena raises questions about DNN representations, as well
as the properties of natural images themselves.

\comment{This phenomenon raises questions about DNN
representations, as well as the properties of natural images themselves.}

\comment{{{
Research on adversarial images to date has focused mainly on disrupting 
classification, i.e., producing images classified with labels that are patently 
inconsistent with human perception. This phenomenon is not entirely 
interesting, because the disrupted image representation in the layer before 
classification only needs to lie within a large and potentially unbounded 
region of the feature space corresponding to the designated erroneous class, 
without requiring the disrupted representation to be like real data, 
i.e., representation of some real natural image.  
Coupled with the fact 
that input space usually has much more degrees of freedom than the final 
representation space (for e.g.\ , $\sim10^{5}$D vs $\sim10^{3}D$ for 
typical ImageNet classification nets), it is not surprising that labels 
could be disrupted easily. Indeed, it has been shown that lack of 
adversarial robustness is not limited to DNNs\citep{GoodfellowEtalICLR2015},
and is a problem for classifiers in 
general \citep{FawziEtalICLR2015,fawzi2015analysis}. Orthogonal to the 
ongoing development of theory on adversarial perturbation for classification, 
in this paper we introduce a more disturbing category of adversarial 
images that are confused with other images, not just in the class label, 
but in the internal representation. In short, this new adversarial 
perturbation phenomenon demonstrates that it is possible to trick a 
DNN to confound almost any image with any other chosen image.
}}}

%% focus directly on properties of representations learned by 
%% deep networks and we introduce
%% What could be more interesting is adversaries 
%% that their internal representations at much higher dimensions and even over 
%% complete projections fools the model, i.e., while the input image changes 
%% almost imperceptively the internal representation resembles of another 
%% natural image.

\comment{{{
Deep neural networks (DNN) have shown pervasive successes in compute vision, from classification\cite{}, detection\cite{}, segmentation\cite{} to action recognition\cite{}, etc. However, some recent discoveries showed that advarsarially designed pixel value perturbation can cause DNNs to misclassify\cite{}; while another work demonstrated that on completely unrecognizable and unnatural images, DNNs can produce high confidence predictions\cite{}. These images specifically designed to trick a given model are called {\em adversarial examples}. There has been conflicting views about the nature and relevance, especially when they are shown to exist for other classifiers as well, rather than specific to DNNs\cite{}. Regardless the truth about adversarial phenomenon on classification labels, in this work, we show the existence of a worse type of adversarial perturbation that can cause DNNs to confuse any natural image with any other given image according to the internal representation, rather than classification output. Because the learned representation is central to modern computer vision systems that rely on DNNs, our work shows that despite the widespread euphoria about recent successes, there is something troubling about the DNN representation of natural images.

%% Conflicting explanations and evidences have been 
%%  about showed that there are small pertuabation to inputs or. Central to the success of DNNs is their ability to learn hierarchical distributed representation from massive amount of data. 
%% In this work,
Specifically, given a source image, a target image, and the DNN, our proposed optimization finds small perturbations to the source image, which can cause the internal representation on a given layer and above to be very similar to that of the target image, and dissimilar to the representation of the source (Fig.\ref{fig:summary} \yanshuai{TBD: make a illustration like the ones we drawn on board about image space and feature space}). Furthermore, we show that the representation of the adversarial example is an inlier with respect to nearby points in the feature space, making it indistinguishable in terms of representation from other natural images in the vicinity of the target, while visually similar to the source for humans. We also demonstrate that the fact about being inlier does not hold for previously proposed adversarial examples for forcing misclassification, making our proposed adversarial examples qualitatively different. Finally, we demonstrate that previous explanations for adversarial examples do not account for our observations.

% Given the widespread use of 
%\yanshuai{To be continued}

%% , which given any pair of source and target natural images, can create small and often imperceptible changes to the source image, but which make the internal representation to be very similar to 
%% For instance, when trained on a large generic natural image datasets such as the ImageNet\cite{}, the representation learned by convolutional neural networks (convnets) has been shown to generalize well to many other tasks and datasets\cite{}. Such pre-trained or fine-tuned deep feature extractors have become the cornerstone in many modern computer vision applications\cite{}. However, despite the euphoria, recent works showed that DNNs can be tricked into producing erroneous prediction labels in bizarre ways when facing adversarial images\cite{}. While such adversarial examples against classification outputs are interesting, it is unclear whether the 
}}}

\section{Related Work}
\label{related}
\vspace*{-0.1cm}

% There are two main approaches to construct adversarial images 
% that disrupt DNN classification.  These approaches differ 
% fundamentally in whether they reveal the behavior of DNNs in 
% the neighborhood of natural images, or the adversarial images 
% themselves are overtly unnatural.

Several methods for generating adversarial images have appeared in 
recent years.  \cite{NguyenEtAlCVPR2015} describe an evolutionary 
algorithm to generate images comprising 2D patterns that are classified 
by DNNs as common objects with high confidence (often $99\%$). While 
interesting, such adversarial images are quite different from the 
natural images used as training data.  Because natural images only 
occupy a small volume of the space of all possible images, it is 
not surprising that discriminative DNNs trained on natural 
images have trouble coping with such out-of-sample data.  

\cite{SzegedyElatICLR2014} focused on adversarial images that appear 
natural. They used gradient-based optimization on the classification 
loss, with respect to the image perturbation, $\epsilon$.  The 
magnitude of the perturbation is penalized ensure that the perturbation
is not perceptually salient.  Given an image $I$, a DNN classifier $f$, 
and an erroneous label $\ell $, they find the perturbation $\epsilon$ that 
minimizes ${loss(f(I+\epsilon), \ell) + c\|\epsilon\|^2}$.
% \begin{equation}
% \phantom{E=}loss(f(I+\epsilon), l) + c\|\epsilon\|^2  ~.
% \end{equation}
Here, $c$ is chosen by line-search to find the smallest $\epsilon$ that 
achieves $f(I+\epsilon) = \ell$. The authors argue that the resulting 
adversarial images occupy low probability ``pockets'' in the manifold,
acting like ``blind spots'' to the DNN. The adversarial construction in 
our paper extends the approach of \cite{SzegedyElatICLR2014}. In 
Sec.~\ref{method}, we use gradient-based optimization to find small image 
perturbations.  But instead of inducing misclassification, we induce dramatic 
changes in the internal DNN representation.  

Later work by \cite{GoodfellowEtalICLR2015} showed that adversarial images 
are more common, and can be found by taking steps in the direction of the 
gradient of $loss(f(I+\epsilon), \ell)$.  \cite{GoodfellowEtalICLR2015} 
also show that adversarial examples exist for other models, including
linear classifiers.  They argue that the problem arises when models 
are ``too linear". \cite{FawziEtalICLR2015} later propose a more general 
framework to explain adversarial images, 
formalizing the intuition that the problem occurs when DNNs and other 
models are not sufficiently ``flexible'' for the given classification task. 

In Sec.~\ref{sec:experiments}, we show that our new category of adversarial 
images exhibits qualitatively different properties from those above.
In particular, the DNN representations of our adversarial images are 
very similar to those of natural images.  They do not appear unnatural 
in any obvious way, except for the fact that they remain inconsistent 
with human perception.

%% \fartash{This is now only a short list of related work that has cited the initial work of 'Intriguing...'}

%% Main related works:

%% Initial work was done in \cite{szegedy2013intriguing} suggesting that given an image, for all possible choices of labels, an adversarial example can be generated that the network condfidently classifies the example as the target class.

%% \cite{nguyen2014deep} Fooling convnets with examples generated by an evolutionary algorithm

%% \cite{goodfellow2014explaining} Linear Perturbation theory

%% \cite{mahendran2014understanding} Inverting convnets

%% \cite{goodfellow2014generative} Generative Adversarial Nets

%% \cite{gu2014towards} Works on solving the problem

%% \cite{fawzifundamental} \cite{fawzi2015analysis} Works on theory

%% \cite{guo2015cnn} We are undermining works like this

%% \cite{wei2015understanding} \cite{tabacof2015exploring} Related work regarding internal representation

\section{Adversarial Image Generation}
\label{method}

Let $I_s$ and $I_g$ denote the {\em source} and {\em guide} images.
Let $\phi_k$ be the mapping from an image to its internal DNN representation 
at layer $k$.  Our goal is to find a new image, $I_\alpha$, such that 
the Euclidian distance between $\phi_k(I_\alpha)$ and $\phi_k(I_g)$ is 
as small as possible, while $I_\alpha$ remains close to the source $I_s$.  
More precisely, $I_\alpha$ is defined to be the solution to a 
constrained optimization problem:
\begin{align}
I_{\alpha} = \arg\min_{I} \,  \| \, \phi_k(I)-\phi_k(I_g) \,\|^{2}_2 & 
\phantom{=E} 
\label{adv_objective} \\
\text{subject to}~~ \| I - I_s \|_{\infty} < \delta &\phantom{=E}  
\label{infnorm_bound}
\end{align}
The constraint on the distance between $I_\alpha$ and $I_s$ is formulated
in terms of the $L_{\infty}$ norm to limit the maximum deviation of any 
single pixel color to $\delta$.  The goal is to constrain the degree 
to which the perturbation is perceptible.  While the $L_\infty$ 
norm is not the best available measure of human visual discriminability 
(e.g., compared to SSIM \citep{WangEtalPAMI2004}), it is superior to 
the $L_2$ norm often used by others.

Rather than optimizing $\delta$ for each image,
% \comment{ as done to find the weight on the perturbation penalty 
% in \cite{SzegedyElatICLR2014},} 
we find that a fixed value of $\delta = 10$ (out of 255) produces 
compelling adversarial images with negligible perceptual distortion. 
Further, it works well with different intermediate layers, different 
networks and most images. 
% This simplifies adversarial generation and analysis.  
We only set $\delta$ larger when optimizing lower layers,
close to the input (e.g., see Fig.\ \ref{fig:delta_layer}).  As $\delta$ 
increases distortion becomes perceptible, but there is little or no perceptible 
trace of the guide image in the distortion.
For numerical optimization,  we use l-BFGS-b, with the inequality 
(\ref{infnorm_bound}) expressed as a box constraint around $I_s$.

%% For notational convenience below we use $\source$, $\guide$, and $\adv$ 
%% to denote source, guide and adversarial images.  We also use $\adv_{ij}^k$ 
%% to denote the DNN representation at layer $k$, from the adversarial image 
%% built from source $i$ and guide $j$.

\makeatletter
\newcommand{\thickhline}{%
    \noalign {\ifnum 0=`}\fi \hrule height 1pt
    \futurelet \reserved@a \@xhline
}
\newcolumntype{"}{@{\hskip\tabcolsep\vrule width 4pt\hskip\tabcolsep}}
\makeatother

\begin{figure*}[t]
\centering
\renewcommand{\arraystretch}{1}
\setlength\tabcolsep{2pt}
\begin{tabular}{ | >{\centering\arraybackslash} m{\dimexpr 0.115\linewidth}
>{\centering\arraybackslash}m{\dimexpr 0.115\linewidth} | 
>{\centering\arraybackslash}m{\dimexpr 0.115\linewidth}
>{\centering\arraybackslash}m{\dimexpr 0.115\linewidth} | 
>{\centering\arraybackslash}m{\dimexpr 0.115\linewidth}
>{\centering\arraybackslash}m{\dimexpr 0.115\linewidth} | 
>{\centering\arraybackslash}m{\dimexpr 0.115\linewidth}  
>{\centering\arraybackslash}m{\dimexpr 0.115\linewidth}@{} | }
\hline
{\footnotesize Source} & 
{\footnotesize Guide} & 
{\footnotesize  $I_{\adv}^{\text{FC}7}\!,\delta\!=\!5$} & 
%{\footnotesize $\Delta_\source(\adv^{\text{FC}7})_5 $} & 
{\footnotesize $\Delta_\source $} & 
{\footnotesize $I_{\adv}^{P5}\!,\delta\!=\!10$} & 
%{\footnotesize $\Delta_\source(\adv^{P5})_{10}$} & 
{\footnotesize $\Delta_\source$} & 
{\footnotesize $I_{\adv}^{C3}\!,\delta\!=\!{15}$} & 
%{\footnotesize $\Delta_\source(\adv^{C3})_{20}$} \\ 
{\footnotesize $\Delta_\source$} \\ 
\hline
\includegraphics[width=\linewidth,height=.75\linewidth]{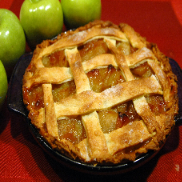} &
\includegraphics[width=\linewidth,height=.75\linewidth]{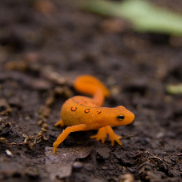} &
\includegraphics[width=\linewidth,height=.75\linewidth]{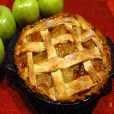} &
\includegraphics[width=\linewidth,height=.75\linewidth]{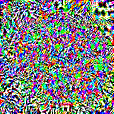} &\iffalse
\includegraphics[width=\linewidth,height=.75\linewidth]{./imgs/pie_t10_g7_fc7/27img.png} &
\includegraphics[width=\linewidth,height=.75\linewidth]{./imgs/pie_t10_g7_fc7/27dif.png} &\fi
\includegraphics[width=\linewidth,height=.75\linewidth]{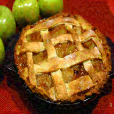} &
\includegraphics[width=\linewidth,height=.75\linewidth]{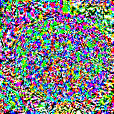} &
\includegraphics[width=\linewidth,height=.75\linewidth]{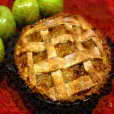} &
\includegraphics[width=\linewidth,height=.75\linewidth]{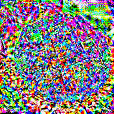} \\
\includegraphics[width=\linewidth,height=.75\linewidth]{./imgs/pie.png} &
\includegraphics[width=\linewidth,height=.75\linewidth]{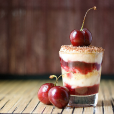} &
\includegraphics[width=\linewidth,height=.75\linewidth]{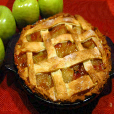} &
\includegraphics[width=\linewidth,height=.75\linewidth]{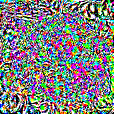} &\iffalse
\includegraphics[width=\linewidth,height=.75\linewidth]{./imgs/pie_t10_g7_fc7/927img.png} &
\includegraphics[width=\linewidth,height=.75\linewidth]{./imgs/pie_t10_g7_fc7/927dif.png} &\fi
\includegraphics[width=\linewidth,height=.75\linewidth]{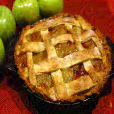} &
\includegraphics[width=\linewidth,height=.75\linewidth]{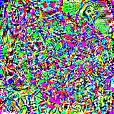} &
\includegraphics[width=\linewidth,height=.75\linewidth]{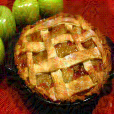} &
\includegraphics[width=\linewidth,height=.75\linewidth]{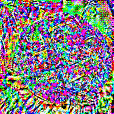} \\
\iftrue
\includegraphics[width=\linewidth,height=.75\linewidth]{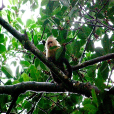} &
\includegraphics[width=\linewidth,height=.75\linewidth]{./imgs/927.png} &
\includegraphics[width=\linewidth,height=.75\linewidth]{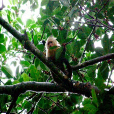} &
\includegraphics[width=\linewidth,height=.75\linewidth]{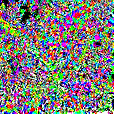} &\iffalse
\includegraphics[width=\linewidth,height=.75\linewidth]{./imgs/train_378_t10_g7_fc7/927img.png} &
\includegraphics[width=\linewidth,height=.75\linewidth]{./imgs/train_378_t10_g7_fc7/927dif.png} &\fi
\includegraphics[width=\linewidth,height=.75\linewidth]{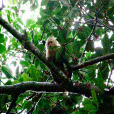} &
\includegraphics[width=\linewidth,height=.75\linewidth]{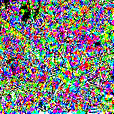} &
\includegraphics[width=\linewidth,height=.75\linewidth]{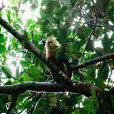} &
\includegraphics[width=\linewidth,height=.75\linewidth]{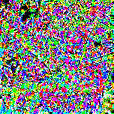} \\
\fi
\hline
\end{tabular}
\caption{Each row shows examples of adversarial images, optimized
using different layers of Caffenet (FC$7$, P$5$, and C$3$), and different 
values of $\delta=(5, 10, 15)$.  Beside each adversarial image is the 
difference between its corresponding source image.}
\label{fig:adv_caffenet}
\vspace*{-0.1cm}
\end{figure*}

Figure~\ref{fig:adv_caffenet} shows nine adversarial images generated 
in this way, all using the well-known BVLC Caffe Reference model 
(Caffenet) \citep{jia2014caffe}.  
Each row in Fig.~\ref{fig:adv_caffenet} shows a source, a guide, and three 
adversarial images along with their differences from the corresponding source.
The adversarial examples were optimized with different perturbation bounds 
($\delta$), and using different layers, namely FC$7$ (fully connected level 7), 
P$5$ (pooling layer 5), and C3 (convolution layer 3).  Inspecting the 
adversarial images, one can see that larger values of $\delta$ allow more 
noticeable perturbations.  That said, we have found no natural images in which 
the guide image is perceptible in the adversarial image.  Nor is there 
a significant amount of salient structure readily visible in the difference 
images.
  
While the class label was not an explicit factor in the optimization, we 
find that class labels assigned to adversarial images by the DNN are almost 
always that of the guide.  For example, we took 100 random source-guide 
pairs of images from Imagenet ILSVRC data \citep{deng2009imagenet}, and 
applied optimization using layer FC7 of Caffenet, with $\delta = 10$.  
We found that class labels assigned to adversarial images were never 
equal to those of source images. Instead, in 95\% of cases they matched 
the guide class.  This remains true for source images from training, 
validation, and test ILSVRC data.

We found a similar pattern of behavior with other networks and datasets, 
including AlexNet \citep{krizhevsky2012imagenet}, GoogleNet 
\citep{szegedy2014going}, and VGG CNN-S \citep{chatfield2014return}, 
all trained on the Imagenet ILSVRC dataset.  We also used AlexNet 
trained on the Places205 dataset, and on a hybrid dataset comprising 205 
scene classes and 977 classes from ImageNet \citep{zhou2014learning}.
In all cases, using 100 random source-guide pairs the class labels 
assigned to the adversarial images do not match the source.  Rather, in 
97\% to 100\% of all cases the predicted class label is that of the guide.

% \david{I have removed most discussion of the adversarial image
% representation being the 1-NN for the guide, since that may be 
% better placed at the beginning of section 4, to add to the results
% in Fig 4 about how much closer the adversarial rep is to the guide
% compared to the source.}

Like other approaches to generating adversarial images 
(e.g., \cite{SzegedyElatICLR2014}), we find that those generated 
on one network are usually misclassified by other networks 
Using the same 100 source-guide pairs with each of the models above,  
we find that, on average, 54\% of adversarial images obtained from one 
network are misclassified by other networks.  That said, they are usually 
not consistently classified with the same label as the guide on different 
netowrks.

%%%%%%%%%%%%%%%%%%%%%%%%%%%%%%%%%%%%%%%%%%%%%%%%%%%%%%%%%%%%%%%%%%%%%%%%
%%%%%%%%%%%%%%%%%%%%%%%%%%%%%%%%%%%%%%%%%%%%%%%%%%%%%%%%%%%%%%%%%%%%%%%%

%%%%%%%%%%
\comment{{{
\begin{table*}[ht] 
\resizebox{\linewidth}{!}{\centering
\begin{tabular}{|c|c|c|c|c|c|} \hline 
Model & Layer & Split sets & Out $\class{\source}$ & 
In $\class{\guide}$ & $\guide=\neigh{1}{\adv}$\\
\hline
CaffeNet&       FC$7$&  train-train&    $100$&  $95$&   $95$\\
AlexNet&        FC$7$&  train-train&    $100$&  $97$&   $96$\\
GoogleNet&      pool5/7x7\_s1&  train-train&    $100$&  $100$&  $100$\\
VGG CNN S&      FC$7$&  train-train&    $100$&  $99$&   $99$\\
Places205 AlexNet&      FC$7$&  train-train&    $100$&  $100$&  $98$\\
Places205 Hybrid&       FC$7$&  train-train&    $100$&  $99$&   $99$\\
Flickr Style&   FC$8$ Flickr&   train-train&    $100$&  $100$&  $95$\\
Flickr Style&   FC$7$&  train-train&    $98$&   $45$&   $32$\\
\hline
CaffeNet&       FC$7$&  train-test&     $100$&  $97$&   $96$\\
AlexNet&        FC$7$&  train-test&     $100$&  $99$&   $97$\\
GoogleNet&      pool5/7x7\_s1&  train-test&     $100$&  $99$&   $99$\\
VGG CNN S&      FC$7$&  train-test&     $100$&  $99$&   $99$\\
Places205 AlexNet&      FC$7$&  train-val&      $100$&  $99$&   $99$\\
Places205 Hybrid&       FC$7$&  train-val&      $100$&  $100$&  $100$\\
Flickr Style&   FC$8$ Flickr&   train-test&     $100$&  $100$&  $95$\\
Flickr Style&   FC$7$&  train-test&     $98$&   $41$&   $31$\\
\hline
CaffeNet&       FC$7$&  val-val&        $100$&  $98$&   $0$\\
\hline
\end{tabular}}
\caption{Results for generating adversarials on various networks.  
For each row, $100$ random samples are drawn from indicated splits 
of the dataset and optimized for $500$ iterations. 
{\it Out $\class{\source}$\/} is the number of adversarials that 
are predicted by the network to have a different label from source.  
{\it In $\class{\guide}$\/} of the adversarials are predicted to 
be from the class of guide.  $\guide=\neigh{1}{\adv}$ means the 
adversarial is so close to the guide that guide is its $1$NN\@.}
\label{tb:generalize_tb1}
\end{table*}
}}}
%%%%%%%%%%

\comment{{{
The Flickr Style has $80,000$ images from Flickr, 
categorized into 20 classes of photo styles.

We repeat the experiments described in Section~\ref{RankAnalysis} on 
other networks to show that the generated adversarial images look like 
inliers in the guide class for various CNN networks. For networks trained 
on Flickr style dataset we show that it is possible to find adversarials 
on the last layer before classification but for layers below, sometimes 
the optimization quickly ends up in local minima.

Table~\ref{tb:generalize_tb1} shows that not only our adversarials almost
always have the label of the guide, but also they are almost always so 
close to the guide that the first nearest neighbor of them in the class 
of the guide is the guide itself. For all networks, except for FC$7$ 
on Flickr, the difference in rank, $\rankdiff{3}$, is better than $-7$ 
which means that for half of the adversarials, we are more similar 
to the guide compared to $93\%$ of other images in that class.

On the fine-tuned model on Flickr Style, we observe that we can quickly 
generate adversarials that we can optimize the FC$8$, the unnormalized 
class scores, to be the same as the guide with $\rankdiff{3}$ being $0$. 
However, when doing the optimization on FC$7$, for $55$ examples on 
train-train split and $59$ examples on train-test split, the optimization 
quickly converges to local minima. The interesting observation is that, 
almost always, the local minima has $\rank{3}=0$. This means that the 
optimization has converged to the most dense regions of a class.

%%%%%%%%%%
\begin{table*}[ht] \resizebox{\linewidth}{!}{\centering
        
\begin{tabular}{|c|c|c|c|} \hline Model & Layer &Same $3$NN & Same
$2/3$ 3NN\\ \hline
CaffeNet&       FC$7$&  $71$&   $24$\\
AlexNet&        FC$7$&  $72$&   $25$\\
GoogleNet&      pool5/7x7\_s1&  $87$&   $13$\\
VGG CNN S&      FC$7$&  $84$&   $16$\\
Places205 AlexNet&      FC$7$&  $91$&   $9$\\
Places205 Hybrid&       FC$7$&  $85$&   $15$\\
Flickr Style&   FC$8$ Flickr&   $100$&  $0$\\
Flickr Style&   FC$7$&  $28$&   $28$\\
\hline
CaffeNet&       FC$7$&  $70$&   $29$\\
AlexNet&        FC$7$&  $79$&   $20$\\
GoogleNet&      pool5/7x7\_s1&  $97$&   $3$\\
VGG CNN S&      FC$7$&  $79$&   $20$\\
Places205 AlexNet&      FC$7$&  $91$&   $9$\\
Places205 Hybrid&       FC$7$&  $84$&   $14$\\
Flickr Style&   FC$8$ Flickr&   $100$&  $0$\\
Flickr Style&   FC$7$&  $25$&   $30$\\
\hline
CaffeNet&       FC$7$&  $78$&   $20$\\
\hline
    \end{tabular}}
  \caption{This table shows that for most of the time the nearest neighbors of
  the guide and the adversarial are exactly the same data points. The first
  column is the number of cases where all 3 nearest neighbors are the same.
  The second column is when 2 out of 3 are the same.}
 \label{tb:generalize_tb1_2}
\end{table*}
}}}

%%%%%%%%%%%%
\comment{{{
\begin{table*}[ht] \resizebox{\linewidth}{!}{\centering
        \begin{tabular}{|c|c|c|c|c|c|c|} \hline Model Name & Layer & Split sets
        & Out $\class{\source}$& In $\class{\guide}$ & $\guide=\neigh{1}{\adv}$
        & $\rankdiff{3}$ median, [min, max]\\ \hline CaffeNet&       FC$7$&
        train-train&    $100$&  $95$&   $95$&   $-5.76, [-54.83, 0.00]$\\
        AlexNet&        FC$7$&  train-train&    $100$&  $97$&   $96$&   $-5.64,
        [-38.39, 0.00]$\\ GoogleNet&      pool5/7x7\_s1&  train-train& $100$&
        $100$&  $100$&  $-1.94, [-12.87, 0.10]$\\ VGG CNN S& FC$7$&
        train-train&    $100$&  $99$&   $99$&   $-3.37, [-26.34,
    0.00]$\\ Places205 AlexNet&      FC$7$&  train-train&    $100$&  $100$&
        $98$&   $-1.25, [-18.20, 8.04]$\\ Places205 Hybrid&       FC$7$&
        train-train&    $100$&  $99$&   $99$&   $-1.30, [-8.96, 8.29]$\\ Flickr
        Style&   FC$8$ Flickr&   train-train&    $100$&  $100$&  $95$&   $0.00,
        [0.00, 0.08]$\\ Flickr Style&   FC$7$&  train-train&    $98$&   $45$&
        $32$&   $-10.28,        [-37.35, -2.39]$\\ \hline CaffeNet&
        FC$7$&  train-test&     $100$&  $97$&   $96$&   $-5.37, [-27.38,
    0.00]$\\ AlexNet&        FC$7$&  train-test&     $100$&  $99$&   $97$&
        $-6.10, [-42.98, 0.23]$\\ GoogleNet&      pool5/7x7\_s1&  train-test&
        $100$&  $99$&   $99$&   $-1.62, [-10.18, 0.51]$\\ VGG CNN S&
        FC$7$&  train-test&     $100$&  $99$&   $99$&   $-3.36, [-18.21,
    0.73]$\\ Places205 AlexNet&      FC$7$&  train-val&      $100$&  $99$&
        $99$&   $-1.16, [-8.30, 3.25]$\\ Places205 Hybrid&       FC$7$&
        train-val&      $100$&  $100$&  $100$&  $-1.38, [-7.39, 5.87]$\\ Flickr
        Style&   FC$8$ Flickr&   train-test&     $100$&  $100$&  $95$&   $0.00,
        [0.00, 0.04]$\\ Flickr Style&   FC$7$&  train-test&     $98$&   $41$&
        $31$&   $-10.94,        [-28.96, 0.00]$\\ \hline CaffeNet&       FC$7$&
        val-val&        $100$&  $98$&   $-$&    $-6.29, [-25.58, -0.08]$\\
        \hline
    \end{tabular}}
  \caption{Results for generating adversarials on various networks.  For each
      row, $100$ random samples are drawn from indicated splits of the dataset
      and optimized for $500$ iterations. {\it Out $\class{\source}$\/} is the
      number of adversarials that are predicted by the network to have
      a different label from source.  {\it In $\class{\guide}$\/} of the
      adversarials are predicted to be from the class of guide.
  $\guide=\neigh{1}{\adv}$ means the adversarial is so close to the guide that
  guide is its $1$NN\@.  Column of $\rankdiff{3}$ shows statistics for the
  difference of ranks.}
 \label{tb:generalize_tb1_prev}
\end{table*}

\begin{table}[]
% \resizebox{\linewidth}{!}{\centering
\begin{center}
\begin{footnotesize}
\begin{tabular}{|c|c|c|c|c|}
\hline
\multicolumn{5}{|c|}{train-train}\\
\hline
Model name      &CaffeNet       &AlexNet        &GoogleNet      &VGG CNN S \\
\hline
CaffeNet &      100,95 &        61,3 &  42,0 &  53,1 \\
AlexNet &       61,0 &  100,97 &        46,0 &  57,0 \\
GoogleNet &     33,1 &  36,0 &  100,100 &       42,0 \\
VGG CNN S &     53,0 &  55,0 &  49,1 &  100,99 \\
            \hline
            \multicolumn{5}{|c|}{train-test}\\
            \hline
Model name      &CaffeNet       &AlexNet        &GoogleNet      &VGG CNN S \\
\hline
CaffeNet &      100,97 &        67,1 &  58,0 &  65,1 \\
AlexNet &       71,1 &  100,99 &        54,0 &  67,0 \\
GoogleNet &     50,0 &  49,1 &  100,99 &        48,0 \\
VGG CNN S &     59,2 &  57,0 &  61,0 &  100,99 \\
            \hline
\end{tabular}
\end{footnotesize}
\end{center}
% }
\caption{Results for generating an adversarial for one network and testing it
on another network. The rows show the network for generating the adversarial
and the column shows the network used for testing the adversarials. For each
row, $100$ adversarials generated for Table~\ref{tb:generalize_tb1} is tested
on networks in the columns. Each cell shows the number of {\it Out
$\class{\source}$\/} and {\it In $\class{\guide}$}. } \label{tb:generalize_tb2}
\end{table}
}}}

\comment{
The adversarial images generated by our optimization are designed to be 
close, if not perceptually equivalent, to a given source image (demonstrated 
in Fig.~\ref{fig:adv_caffenet}), with an internal represetation close to 
the guide. 
}

%%%%%%%%%%%%%%%%%%%%%%%%%%%%%%%%%%%%%%%%%%%%%%%%%%%%%%%%%%%%%%%%%%%%%%%%
%%%%%%%%%%%%%%%%%%%%%%%%%%%%%%%%%%%%%%%%%%%%%%%%%%%%%%%%%%%%%%%%%%%%%%%%

\begin{figure*}[t!]
\centering
\begin{subfigure}[t]{\linewidth}
{
\renewcommand{\arraystretch}{1}
\setlength\tabcolsep{2pt}
\begin{tabular}{|
>{\centering\arraybackslash}m{0.09\linewidth} |
>{\centering\arraybackslash}m{0.167\linewidth} |
>{\centering\arraybackslash}m{0.167\linewidth}
>{\centering\arraybackslash}m{0.167\linewidth}
>{\centering\arraybackslash}m{0.167\linewidth} |
>{\centering\arraybackslash}m{0.167\linewidth} | }
\hline
 & Source & $\text{FC}7$ & $\text{P}5$ & C$3$ &Guide  \\\hline  
Input & 
%\begin{tikzonimage}[width=\linewidth, 
%height=.75\linewidth]{./imgs/train_730.png}[image label]
%\node{S}; \end{tikzonimage}
\includegraphics[width=\linewidth,height=.75\linewidth]{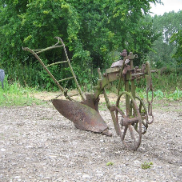} 
%& \begin{tikzonimage}[width=\linewidth, height=.75\linewidth]
%    {./imgs/train_730_t10_fc7_100043mat/orig.png}[image label] 
%\node{F};\end{tikzonimage}
 &
 \includegraphics[width=\linewidth,height=.75\linewidth]{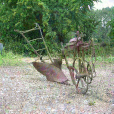} 
 &
% \begin{tikzonimage}[width=\linewidth, height=.75\linewidth]
%    {./imgs/train_730_t10_pool5_100043mat/orig.png}[image label] 
%\node{P};\end{tikzonimage}
\includegraphics[width=\linewidth,height=.75\linewidth]{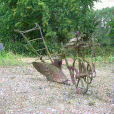} 
&
% \begin{tikzonimage}[width=\linewidth, height=.75\linewidth]
%    {./imgs/train_730_t20_conv3_100043mat/orig.png}[image label] 
%\node{C};\end{tikzonimage}
\includegraphics[width=\linewidth,height=.75\linewidth]{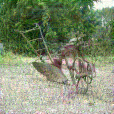} 
&
% \begin{tikzonimage}[width=\linewidth, height=.75\linewidth]
%    {./imgs/val_43.png}[image label] \node{G};\end{tikzonimage} \\
\includegraphics[width=\linewidth,height=.75\linewidth]{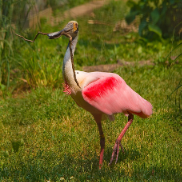} \\
Inv(C$3$) & 
\includegraphics[width=\linewidth,height=.75\linewidth]{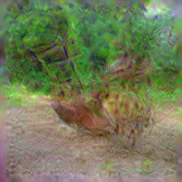} & 
\includegraphics[width=\linewidth,height=.75\linewidth]{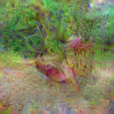} &
\includegraphics[width=\linewidth,height=.75\linewidth]{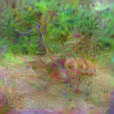} &
\includegraphics[width=\linewidth,height=.75\linewidth]{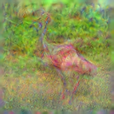} &
\includegraphics[width=\linewidth,height=.75\linewidth]{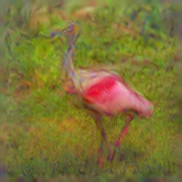} 
\\
Inv($\text{P}5$) & 
\includegraphics[width=\linewidth,height=.75\linewidth]{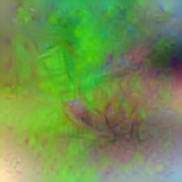} & 
\includegraphics[width=\linewidth,height=.75\linewidth]{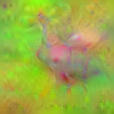} &
\includegraphics[width=\linewidth,height=.75\linewidth]{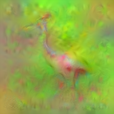} &
\includegraphics[width=\linewidth,height=.75\linewidth]{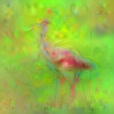} &
\includegraphics[width=\linewidth,height=.75\linewidth]{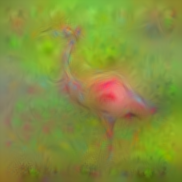} 
\\
Inv($\text{FC}7$) & 
\includegraphics[width=\linewidth,height=.75\linewidth]{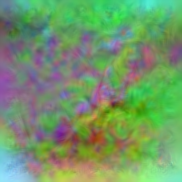} & 
\includegraphics[width=\linewidth,height=.75\linewidth]{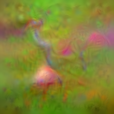} &
\includegraphics[width=\linewidth,height=.75\linewidth]{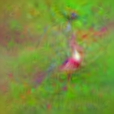} &
\includegraphics[width=\linewidth,height=.75\linewidth]{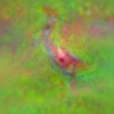} &
\includegraphics[width=\linewidth,height=.75\linewidth]{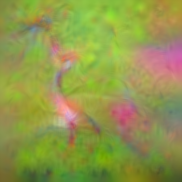} 
\\

\hline
\end{tabular}
}
\end{subfigure}

\vspace*{0.2cm}

\begin{subfigure}[t]{\linewidth}
{
\centering
\renewcommand{\arraystretch}{1}
\setlength\tabcolsep{.1pt}
\begin{tabular}{
|>{\centering\arraybackslash}m{0.205\linewidth}
>{\centering\arraybackslash}m{0.205\linewidth}
>{\centering\arraybackslash}m{0.205\linewidth}|
>{\centering\arraybackslash}m{0.125\linewidth}
>{\centering\arraybackslash}m{0.125\linewidth}
>{\centering\arraybackslash}m{0.125\linewidth}|
}
\hline
% \begin{tikzonimage}[width=\linewidth]
%     {./imgs/f7_730.png}[image label] \node{S}; \end{tikzonimage}&
\includegraphics[width=\linewidth]{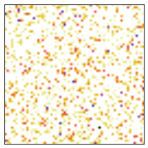} &
% \begin{tikzonimage}[width=\linewidth]
%     {./imgs/f7_730_43.png}[image label] \node{F}; \end{tikzonimage}&
\includegraphics[width=\linewidth]{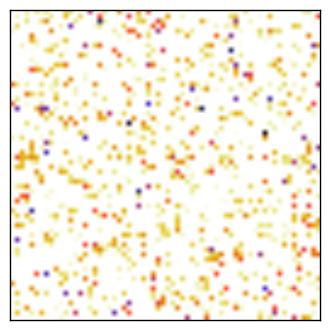} &
% \begin{tikzonimage}[width=\linewidth]
%     {./imgs/f7_43.png}[image label] \node{G}; \end{tikzonimage}&
\includegraphics[width=\linewidth]{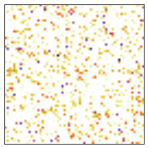} &
% \begin{tikzonimage}[height=\linewidth, angle=90]
%     {./imgs/p5_730.png}[image label] \node{S}; \end{tikzonimage}&
\includegraphics[height=\linewidth, angle=90]{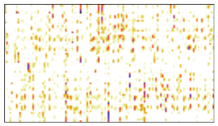} &
% \begin{tikzonimage}[height=\linewidth, angle=90]
% {./imgs/p5_730_43.png}[image label] \node{P};\end{tikzonimage}&
\includegraphics[height=\linewidth, angle=90]{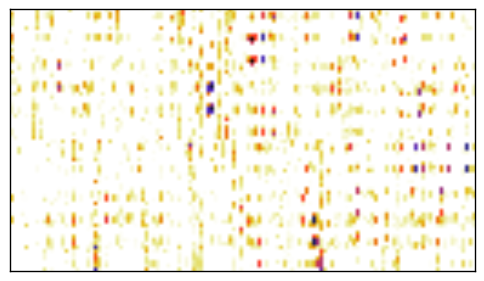} &
% \begin{tikzonimage}[height=\linewidth, angle=90]
%    {./imgs/p5_43.png}[image label] \node{G}; \end{tikzonimage}\\
\includegraphics[height=\linewidth, angle=90]{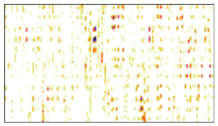}\\
Source & FC7 Advers. & Guide & Source & P5 Advers. & Guide \\ \hline
%\multicolumn{3}{|c|}{FC$7$ Activations} & \multicolumn{3}{c|}{ P$5$ 
%Activations  } \\
%\hline
\end{tabular}
}
\end{subfigure}
\caption{
(Top Panel) The top row shows a source (left), a guide (right), and 
three adversarial images (middle), optimized using layers FC$7$, P$5$, 
and C3 of Caffenet.  The next three rows show images obtained by 
inverting the DNN mapping, from layers C3, P$5$, and FC$7$ 
respectively \citep{MahendranVedaldiCVPR2015}.
(Lower Panel) Activation patterns are shown at layer FC7 for the source,
guide and FC7 adversarial above, and at layer P5 for the source, guide 
and P5 adversarial image above.
% One can see how the internal representations begin to mimic the guide, 
% beginning at the layer used in the optimization.
}
\label{fig:adv_invert}
\end{figure*}

We next turn to consider internal representations -- do they resemble 
those of the source, the guide, or some combination of the two?  
One way to probe the internal representations, following 
\cite{MahendranVedaldiCVPR2015}, is to invert the mapping, thereby 
reconstructing images from internal representations at specific layers. 
The top panel in Fig.~\ref{fig:adv_invert} shows reconstructed images 
for a source-guide pair.  The {\em Input} row displays a source (left), 
a guide (right) and adervarisal images optimized to match representations at 
layers FC7, P5 and C3 of Caffenet (middle).  Subsequent rows show 
reconstructions from the internal representations of these five 
images, again from layers C3, P5 and FC7.
Note how lower layers bear more similarity to the source, while higher
layers resemble the guide.
When optimized using C3, the reconstructions 
from C3 shows a mixture of source and guide.  
In almost all cases we find that internal representations begin 
to mimic the guide at the layer targeted by the optimization.  
These reconstructions suggest that human perception and the DNN 
representations of these adversarial images are clearly at odds with 
one another.

The bottom panel of Fig.~\ref{fig:adv_invert} depicts FC7 and P5 activation 
patterns for the source and guide images in Fig.~\ref{fig:adv_invert}, 
along with those for their corresponding adversarial images.
We note that the adversarial activations are sparse and much more 
closely resemble the guide encoding than the source encoding.
The supplementary material includes several more examples of 
adversarial images, their activation patterns, and reconstructions 
from intermediate layers.

\comment{In particular, with random noise as an initial guess, they optimize
the following regularized objective using stochastic gradient descent
in order to find a reconstructed image $I_*$ with unit $L_2$ norm:
\begin{equation} 
\label{eq:inv_opt}
\frac{\lVert \Phi(\sigma I_*)-\Phi(I)\rVert^2_2 }{ \lVert \Phi(I) \rVert^2_2}
    +\lambda_\alpha\mathcal{R}_\alpha(I_*)+\lambda_{V^\beta}\mathcal{R}_{V^\beta}( I_* )
\end{equation}
where the two regularizers, $\mathcal{R}_\alpha(x)$ and 
$\mathcal{R}_{V^\beta}$, serve to produce natural images with a 
consistent dynamic range. Some form of regularization here is clearly
necessary because the DNN mapping is many-to-one.
}

%%%%%%%%%%%%%%%%%%%%%%%%%%%%%%%%%%%%%%%%%%%%%%%%%%%%%%%%%%%%%%%%%%%%%%%%
%%%%%%%%%%%%%%%%%%%%%%%%%%%%%%%%%%%%%%%%%%%%%%%%%%%%%%%%%%%%%%%%%%%%%%%%
\comment{{{
}}}
%%%%%%%%%%%%%%%%%%%%%%%%%%%%%%%%%%%%%%%%%%%%%%%%%%%%%%%%%%%%%%%%%%%%%%%%
%%%%%%%%%%%%%%%%%%%%%%%%%%%%%%%%%%%%%%%%%%%%%%%%%%%%%%%%%%%%%%%%%%%%%%%%

\section{Experimental Evaluation}
\label{sec:experiments}
\vspace*{-0.2cm}

% \subsubsection{Euclidean distance and nearest neighbors}
% \label{NNAnalysis}
% \subsection{Manifold local tangent space}
% \label{lpca}

We investigate further properties of adversarial images by asking two 
questions. To what extent do internal representations of adversarial 
images resemble those of the respective guides, and are the 
representations unnatural in any obvious way?
To answer these questions we focus mainly on Caffenet, with random 
pairs of source-guide images drawn from the ImageNet ILSVRC datasets. 
% Further experiments in  Sec.\ \ref{sec:closeness} report results on 
% other well-known convolutional neural networks (CNNs).

\comment{{{
To answer these questions quantitatively, we use several
measures of similarity described below in order to show that the adversarial 
images we obtain have tested properties of natural image representations.
They do not appear to be outliers from the training corpus in any significant
way, as judged by the internal representations.
}}}

% The set of sources comprise 20 images drawn at random ILSVRC training, 
% testing and validation sets. For the guide set, for each of 1000 classes 
% in ILSVRC, we draw three images from training and validation sets whoses 
% labels are correctly predicted by Caffenet (for simplicity in quantifying 
% classification behaviour of adversarial images).  The experiments comprise 
% all possible source-guide combinations from these sets.  

% We start by inverting the reprsentation of the adversarials, following
% \cite{MahendranVedaldiCVPR2015}, to show that although the purturbation is
% impreceptible, but the inverted representation of the CNN looks almost 
% like the guide image.  
% To answer the question regarding how natural an image representation looks,
% we model the local
% neighborhoods by various methods and show that not only the representation of
% the adversarial is very close to the guide, but also it is an inlier in that
% local neighborhood.
% In what follows we demontrate the generation of adversarial images
% and quantify many of their interesting properties, some of which 
% are evident above.  

\iffalse
\input{sample_data}
\fi

%%%%%%%%%%%%%%%%%%%%%%%%%%%%%%%%%
\comment{{{\begin{table}[]
\centering
\resizebox{\linewidth}{!}{%
\begin{tabular}{|c|c|c|c|c|c|}
\hline
 Data Type & Imagenet Training & Imagenet Validation & Imagenet Test & Wikipedia & Total \\ \hline
Source\# & $5$ (R) & $5$ (R) + $3$ (M) & $5$ (R) & $3$ (M) & $20$ \\
Guide\# & $3300$ (PR) & $30$(PR) & $100$(R) & $0$ & $3430$ \\ \hline
\end{tabular}
}
\caption{Number of images from each data type present in each data set.
R, PR and M stands for random, partially random and manual selection strategy.}
\label{data_table}
\end{table}
}}}
%%%%%%%%%%%%%%%%%%%%%%%%%%%%%%%%%
%This yields 6660 pairs, in total, across the
%two disjoint guide sets.
% The majority of our experimental results use the BVLC Caffe Refernce 
% model (caffenet).  A list of the layers of caffenet along with some 
% characteristics is given in Table~\ref{tab:caffenet}.  
\comment{
\begin{table}[t]
\centering
\begin{tabular}{|m{0.45\linewidth}|m{0.15\linewidth}m{0.15\linewidth}m{0.15\linewidth}|}
    \hline
    Layer Name   & Pool 5 & FC 6 & FC 7 \\ \hline
    Dimension    & 9216 & 4096 & 4096 \\ \hline
    Mean of $avg_i$ & 1568.683    & 309.848    & 89.859   \\ \hline
    Maximum of $avg_i$ & 1904.253    & 418.522    & 115.334    \\ \hline
    Minimum of $avg_i$ & 1011.284    & 227.786    & 68.805    \\ \hline
    Average over all classes' maximum pairwise distance  & 2271.464    & 565.371    & 157.112    \\ \hline
\end{tabular}
\caption{Caffenet layer properties.
\david{move contents of table into the text in Section 4.2}}
\label{tab:caffenet}
\end{table}
}

%\david{move this.} \yanshuai{Rephrased it, so maybe it's ok to put here 
%now...}
%\sara{moved $\alpha$ source... to 2nd par 4.1}
\comment{
As all analyses henceforth focus on the representation space rather than pixel 
space, for notational convenience, we use $\adv$, $\source$ and $\guide$ to 
denote DNN representations of the source, guide and adversarial images whenever 
there is no confusion about the layer of the representations.
%\sara{moved $a_{ij}$}
}
\comment{
Where appropriate, we use $\adv_{ij}^k$ to denote the DNN representation at 
layer $k$, from the adversarial image built from source $i$ and guide $j$.
}

\vspace*{-0.1cm}
\subsection{Similarity to the Guide representation}
\vspace*{-0.2cm}
\label{sec:closeness}

We first report quantitative measures of proximity between the source, 
guide, and adversarial image encodings at intermediate layers.
Surprisingly, despite the constraint that forces adversarial and source
images to remain perceptually indistinguishable, the intermediate 
representations of the adversarial images are much closer to guides 
than source images. 
More interestingly, the adversarial representations are often nearest 
neighbors of their respective guides. We find this is true for a 
remarkably wide range of natural images.  

% To this end we report similarities between source, guide and adversarial 
% representations, in terms of common, interpretable distances in the 
% feature space, on a large collection of source-guide pairs.  

For optimizations at layer FC7, we test on a dataset comprising over 
20,000 source-guide pairs, sampled from training, test and validation sets 
of ILSVRC, plus some images from Wikipedia to increase diversity. 
For layers with higher dimensionality (e.g., P5), for computational 
expedience, we use a smaller set of 2,000 pairs.  Additional details about 
how images are sampled can be found in the supplementary material.
To simplify the exposition in what follows, we use $\source$, $\guide$ and 
$\adv$  to denote DNN representations of source, guide and adversarial images, 
whenever there is no confusion about the layer of the representations.

\begin{figure*}[t!]
\centering
\comment{\begin{subfigure}[t]{.26\linewidth}
\includegraphics[width=\linewidth]{./imgs/guide_init_all_p5.png}
\caption{\footnotesize{$\left. d(\adv,\!\guide) \middle/ d(\source,\!\guide) \right.$, P5}}
\label{fig:Guide_Init_p5}
\end{subfigure}
\begin{subfigure}[t]{.24\linewidth}
    \includegraphics[width=\linewidth, height=.8\linewidth] 
    {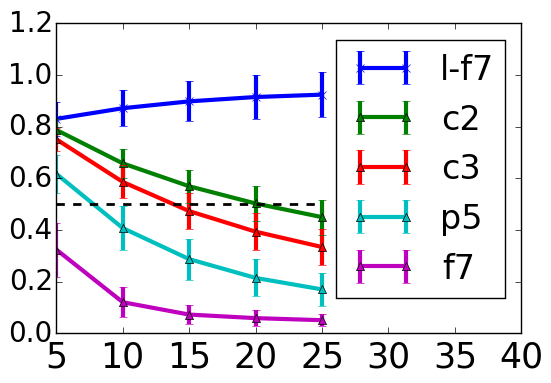}
\caption{\footnotesize{$\left. d(\adv,\!\guide) \middle/ d(\source,\!\guide) 
\right.$, $\delta$}}
\end{subfigure}}
\begin{subfigure}[t]{.24\linewidth}
\includegraphics[width=\linewidth, 
height=.78\linewidth]{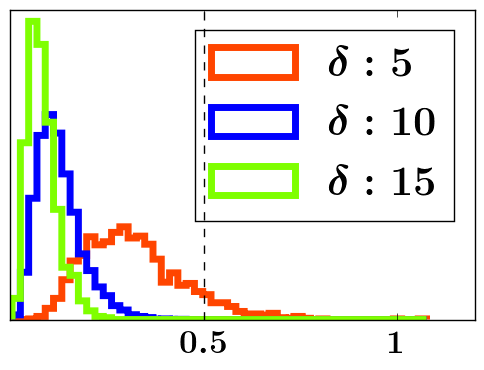}
\caption{\footnotesize{$\left. d(\adv,\!\guide) \middle/ d(\source,\!\guide) \right.$}}
\label{fig:Guide_Init}
\end{subfigure}
\begin{subfigure}[t]{.24\linewidth}
\includegraphics[width=\linewidth, 
height=.78\linewidth]{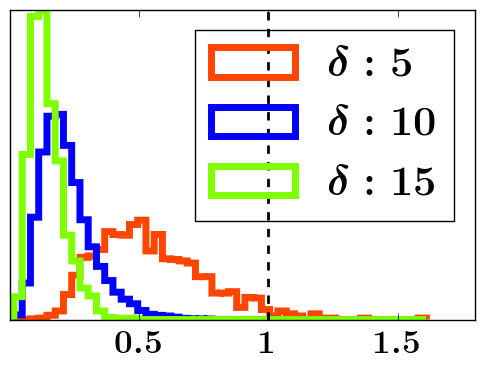}
\caption{\footnotesize{$\left. {d(\adv,\!\guide)} \middle/ {\,\overline{d_1}(\guide)} \right.$}}
\label{fig:guide_nneigh}
\end{subfigure}
\begin{subfigure}[t]{.24\linewidth}
\includegraphics[width=\linewidth, 
height=.78\linewidth]{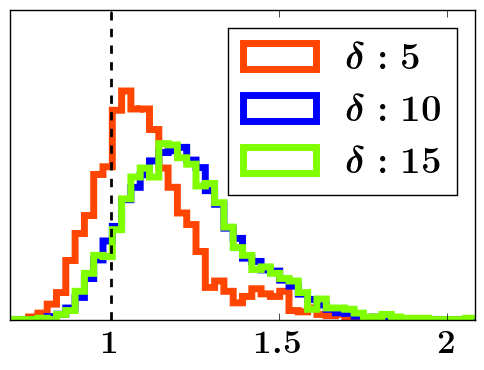}
\caption{\footnotesize{
$\left. {d(\adv,\!\source)} \middle/ {\, \overline{d}(\source)} \right.$}}
\label{fig:Source_Intra}
\end{subfigure}
\vspace*{-0.15cm}
\caption{Histogram of the Euclidean distances between FC7 adversarial encodings 
($\adv$) and corresponding source ($\source$) and  guide  ($\guide$),
for optimizations targetting FC7.
Here, $d(x,y)$ is the distance between $x$ and $y$, $\overline{d}(\source)$ 
denotes the average pairwise distances between points from images of the same
class as the source, and $\overline{d_1}(\guide)$ is the average distance 
to the nearest neighbor encoding among images with the same class as the guide.
Histograms aggregate over all source-guide pairs.}
\label{fig:H_Dist}
\end{figure*}

\vspace*{-0.2cm}
\paragraph{Euclidean Distance:}
As a means of quantifying the qualitative results in Fig.\ 
\ref{fig:adv_invert}, for a large ensemble of source-guide pairs,
all optimized at layer FC7, Fig.~\ref{fig:H_Dist}\subref{fig:Guide_Init} 
shows a histogram of the ratio of Euclidean distance between adversarial 
$\adv$ and guide $\guide$ in FC7, to the distance between source 
$\source$ and guide $\guide$ in FC7.
Ratios less than 0.5 indicate that the adversarial FC7
encoding is closer to $\guide$ than $\source$. 
While one might think that the $L_\infty$ norm constraint on the 
perturbation will limit the extent to which adversarial encodings 
can deviate from the source, we find that the optimization fails 
to reduce the FC7 distance ratio to less than $0.8$ in only $0.1\%$ 
of pairs when $\delta=5$. 
Figure \ref{fig:delta_layer} below shows that if we relax the 
$L_\infty$ bound on the deviation from the source image, then $\alpha$ 
is even closer to $\guide$, and that adversarial encodings become 
closer to $\guide$ as one goes from low to higher layers of a DNN.

% The distance between internal FC$7$ representations of source and guide pairs
% is $110.24$, while the average FC$7$ distances between adversarial with
% $\delta=10$ and guide images is just $13.82$.  between internal FC$7$
% representations of sowurce and guide pairs is $110.24$, while the average
% FC$7$ distances between adversarial with $\delta=10$ and guide images 
% Other histograms in Fig.~\ref{fig:H_Dist} show the proximity of adversarial
% encodings to $\guide$ relative to distances between $\guide$ and its 
% FC7 nearest neighbors from among ILSVRC images.
% 
% Further evaluations show that with $\delta=10$, on average the FC7 
% distance between $\adv$ and $\guide$ is 78\% smaller than the average 
% FC7 distance between other images' representations and their NNs. 

Figure~\ref{fig:H_Dist}\subref{fig:guide_nneigh} compares the FC$7$ 
distances between $\adv$ and $\guide$ to the average FC7 distance between 
representations of all ILSVRC training images from the same class as 
the guide and their FC7 nearest neighbors (NN). 
Not only is $\adv$ often the 1-NN of $\guide$, but the distance between 
$\adv$ and $\guide$ is much smaller than the distance between other 
points and their NN in the same class.
Fig.~\ref{fig:H_Dist}\subref{fig:Source_Intra} shows that the FC7 distance 
between $\adv$ and $\source$ is relatively large compared to typical 
pairwise distances between FC7 encodings of images of the source class.
Only $8\%$ of adversarial images  (at $\delta=10$) are closer to their 
source than the average pairwise FC7 distance within the source class.

% \yanshuai{The rest of this section is still very verbose.}
% \subsubsection{Rank of Average Nearest Neighbors Measure}

% \subsection{Distribution of Average Nearest Neighbor Ranks}
\label{NNAnalysis}

\vspace{-0.2cm}
\paragraph{Intersection and Average Distance to Nearest Neighbors:}
Looking at one's nearest neighbors provides another measure of similarity.
It is useful when densities of points changes significantly through 
feature space, in which case Euclidean distance may be less meaningful.
To this end we quantify similarity through rank statistics on near neighbors.
We take the average distance to a point's $K$ NNs as a scalar score 
for the point.  We then rank that point along with all other points of 
the same label class within the training set.  As such, the rank is a 
non-parametric transformation of average distance, but independant of 
the unit of distance.  We denote the rank of a point $x$ as $\rank{K}{x}$;
we use $K=3$ below.  Since $\adv$ is close to $\guide$ by construction, 
we exclude $\guide$ when finding NNs for adversarial points $\adv$.

\comment{
We also consider rank statistics as a non-parametric way to quantify
similarity between internal representations of adversarial and guide images.
Along with the NN intersections, we compute the average distance to top $K$ NN. 
We consider rank statistics as way of quantifying similarity between average 
distances to the top NNs.  }

Table~\ref{tb:3nn} shows 3NN intersection as well as the difference in rank 
between adversarial and guide encodings, $\rankdiff{3}(\adv, 
\guide)=\rank{3}{\adv}-\rank{3}{\guide}$. When $\adv$ is close enough to 
$\guide$, we expect the intersection to be high, and rank differences to be 
small in magnitude. As shown in Table~\ref{tb:3nn}, in most cases they share 
exactly the same 3NN; and in at least $50\%$ of cases their rank is more 
similar than $90\%$ of data points in that class.  These results are for 
sources and guides taken from the ILSVRC training set.  The same statistics 
are observed for data from test or validation sets.

\comment{Beside networks trained on ImageNet and Places205 dataset, we also 
test a network trained on Flickr Style dataset \citep{karayev2013recognizing} 
which has $80,000$ images categorized into $20$ categories of photography 
styles.  }

\begin{table*}[t] \resizebox{\linewidth}{!}{\centering
        \begin{tabular}{|c|c|c|c|c|}
            \hline Model & Layer & $\cap 3$NN $=3$ & $\cap 3$NN $\geq 2$
            &  $\rankdiff{3}$ median, [min, max] ($\%$)\\
            \hline
CaffeNet~\citep{jia2014caffe}&       FC$7$&  $71$&   $95$&   $-5.98, [-64.69, 0.00]$\\
AlexNet~\citep{krizhevsky2012imagenet}&        FC$7$&  $72$&   $97$&   $-5.64, [-38.39, 0.00]$\\
GoogleNet~\citep{szegedy2014going}&      pool5/$7\times7$\_s1&   $87$&   $100$&  $-1.94, [-12.87, 0.10]$\\
VGG CNN S~\citep{chatfield2014return}&      FC$7$&  $84$&   $100$&  $-3.34, [-26.34, 0.00]$\\
Places205 AlexNet~\citep{zhou2014learning}&      FC$7$&  $91$&   $100$&  $-1.24, [-18.20, 8.04]$\\
Places205 Hybrid~\citep{zhou2014learning}&       FC$7$&  $85$&   $100$&  $-1.25, [-8.96, 8.29]$\\
% Flickr Style~\citep{xiafine}&   FC$8$ Flickr&   $100$&  $100$&  $0.00,  
            % [0.00, 
            %0.08]$\\
% Flickr Style~\cite{xiafine}&   FC$7$&  $28$&   $56$&   $-21.44,        [-98.40, -0.11]$\\
\comment{%
\hline
CaffeNet&       FC$7$&  $70$&   $99$&   $-5.32, [-40.41, 0.00]$\\
AlexNet&        FC$7$&  $79$&   $99$&   $-5.92, [-42.98, 0.23]$\\
GoogleNet&      pool5/$7\times7$\_s1&   $97$&   $100$&  $-1.62, [-10.18, 0.51]$\\
VGG CNN S&      FC$7$&  $79$&   $99$&   $-3.44, [-59.66, 0.73]$\\
Places205 AlexNet&      FC$7$&  $91$&   $100$&  $-1.18, [-8.30, 3.25]$\\
Places205 Hybrid&       FC$7$&  $84$&   $98$&   $-1.38, [-7.39, 5.87]$\\
Flickr Style&   FC$8$ Flickr&   $100$&  $100$&  $0.00,  [0.00, 0.04]$\\
Flickr Style&   FC$7$&  $25$&   $55$&   $-29.16,        [-99.12, 8.68]$\\
\hline
CaffeNet&       FC$7$&  $78$&   $98$&   $-6.29, [-25.58, -0.08]$\\
}
\hline
\end{tabular}} 
\vspace*{-0.1cm}
\caption{Results for comparison of nearest neighbors of the
adversarial and guide.  We randomly select $100$ pairs of guide and
source images such that the guide is classified correctly and the source is
classified to a different class. The optimization is done for a maximum of
$500$ iterations, with $\delta=10$. The statistics are in percentiles.
}
\label{tb:3nn}
\vspace*{-0.1cm}
\end{table*}

\comment{ We have observed a peculiar case for Flickr FC$7$, where the
    optimization very quickly converges to local minima. At the same time, we
    are $100\%$ successful in making the representation exactly the same at
    FC$8$.  Interestingly, at the local minima we see that the
    $\rank{K}{\neigh{1}{\adv}}$ is almost zero which is indicative of a dense
    region where movement to any direction will make the representation farther
    from the guide. If we only look at the adversarials that have come close to
    the guide such that their $1$NN is the guide, the statistics for
$\rankdiff{3}$ is much better, $-10.28, [-37.35, -2.39]$.}

\comment{%

This means that the classification boundaries learned by the network in this
representation is linear and in local neighborhoods, euclidean distance is
expected to be meaningful.

For each guide-source pair, we take the training data that is correctly
classified to the class of the guide and we call it the true-positives of the
guide class. For all the true-positives of the guide class and the adversarial,
we compute the average of their distance to their $3$-nearest-neighbors. We use
this to measure the structure of local neighborhoods around datapoints. For an
adversarial to be an inlier, it needs to be on the underlying manifold of the
guide class and therefore its average distance to its $3$-nearest-neighbors
should come from the same distribution. Also, if the adversarial is close
enough to the guide, it should have the same nearest-neighbors as the guide and
almost the same distance to them as the guide.

Figure~\ref{fig:rank_diff} shows the distribution of difference of ranks
between adversarial and the corresponding guide. This figure shows that the
distribution is concentrated around zero. As shown by the lines indicating the
median of the distribution, we see that for at least $50\%$ of the
adversarials, the difference in the rank with the guide is less than $5\%$ of
the ranks in the class of the guide. This shows that our adversarials, very
often enter a region where the density is very similar to the region of the
guide. This observation besides the closeness of the adversarial to the guide
shows that we are so close to the guide that to the nearest neighbors, there is
little difference between the guide and the adversarial.

\begin{figure}[h] \centering
    \includegraphics[width=.5\linewidth]{./imgs/rank_diff.png} \caption{For
        Caffenet, this shows the distribution of differences between the rank
        an adversarial and its guide, $\rankdiff{3}$. Different plots
        correspond to the split of the dataset that the source and guide are
    selected from.}\label{fig:rank_diff}
\end{figure}

}

\subsection{Similarity to Natural representations}
\label{sec:natural_reps}
\vspace*{-0.2cm}

Having established that internal representations of adversarial 
images ($\adv$) are close to those of guides ($\guide$), we then ask,
to what extent are they typical of natural images? 
That is, in the vicinity of $\guide$, is $\adv$ an inlier, 
with the same characteristics as other points in the neighborhood? 
We answer this question by examining two neighborhood properties: 
1) a probabilistic parametric measure giving the log likelihood of a 
point relative to the local manifold at $\guide$; 2) a geometric 
non-parametric measure inspired by high dimensional outlier detection methods.

For the analysis that follows, let $\Neighs{K}{x}$ denote the set of 
$K$ NNs of point $x$.  Also, let $N_{ref}$ be a set of reference points 
comprising $15$ random points from $\Neighs{20}{\guide}$, and let $N_{c}$ 
be the remaining ``close'' NNs of the guide, $N_{c} = \Neighs{20}{\guide} 
\setminus N_{ref}$.  Finally, let $N_f = \Neighs{50}{\guide} \setminus 
\Neighs{40}{\guide}$ be the set of ``far'' NNs of the guide.
The reference set $N_{ref}$ is used for measurement construction, while
$\adv$, $N_c$ and $N_f$ are scored relative to $\guide$
by the two measures mentioned above. Because we use up to $50$ NNs, for 
which Euclidean distance might not be meaningful similarity measure for 
points in a high-dimensional space like P5, we use cosine distance for 
defining NNs.
(The source images used below are the same $20$ used in 
Sec.\ \ref{sec:closeness}. For expedience, the guide set is a smaller version 
of that used in Sec.\ \ref{sec:closeness}, comprising three images from each 
of only $30$ random classes.)

% \subsubsection{Manifold tangent space}
% \label{sec:tangent_space}
\vspace*{-0.2cm}
\paragraph{Manifold Tangent Space:}
We build a probabilistic subspace model with probabilistic PCA (PPCA) 
around $\guide$ and compare the likelihood of $\adv$ to other points. 
More precisely, PPCA is applied to $N_{ref}$, whose principal space is 
a secant plane that has approximately the same normal direction as the 
tangent plane, but generally does not pass through $\guide$ because of 
the curvature of the manifold. We correct this small offset by shifting the 
plane to pass through $\guide$; with PPCA this is achieved by moving 
the mean of the high-dimensional Gaussian to $\guide$.  We then evaluate 
the log likelihood of points under the model, relative to the log likelihood 
of $\guide$, denoted $\dlike{\cdot}{\guide}=L({\cdot}) - L({\guide})$. 
We repeat this measurement for a large number of guide and source pairs, 
and compare the distribution of $\Delta L$ for $\adv$ with points in
$N_c$ and $N_f$.

For guide images sampled from ILSVRC training and validation sets, 
results for FC7 and P5 are shown in the first two columns of Fig.\ 
\ref{fig:local_inlier}. Since the Gaussian is centred at $\guide$, $\Delta L$
is bounded above by zero. The plots show that $\adv$ is well explained 
locally by the manifold tangent plane. Comparing $\adv$ obtained when 
$\guide$ is sampled from training or validation sets 
(Fig.\ \ref{sf:dl_fc7_train} vs \ref{sf:dl_fc7_val}, \ref{sf:dl_pool5_train} 
vs \ref{sf:dl_pool5_val}), we observe patterns very similar to those in 
plots of the log likelihood under the local subspace models.  This suggests 
that the phenomenon of adversarial perturbation in Eqn.~(\ref{adv_objective}) 
is an intrinsic property of the representation itself,
rather than the generalization of the model.

% \subsubsection{Angular consistency measure}
% \label{sec:angular_const}
\vspace*{-0.2cm}
\paragraph{Angular Consistency Measure:}
If the NNs of $\guide$ are sparse in the high-dimensional feature space, or 
the manifold has high curvature, a linear Gaussian model will be a poor fit.  
So we consider a way to test whether $\adv$ is an inlier in the 
vicinity of $\guide$ that does not rely on a manifold assumption. 
We take a set of reference points near a $\guide$, $N_{ref}$, and measure 
directions from $\guide$ to each point.  We then compare the directions 
from $\guide$ with those from $\adv$ and other nearby points, 
e.g., in $N_{c}$ or $N_{f}$, to see whether $\adv$ is similar to 
other points around $\guide$ in terms of {\em angular consistency}. 
Compared to points within the local manifold, a point far from the 
manifold will tend to exhibit a narrower range of directions to others 
points in the manifold.
Specifically, given reference set $N_{ref}$, with cardinality $k$, and with $z$ 
being $\adv$ or a point from $N_c$ or $N_f$,
our angular consistency measure is defined as
\begin{equation}
\Omega(z,\guide) \, = \, 
\frac{1}{k}
\sum_{ x_i \in N_{ref} }
\frac{\langle x_i - z ,\, x_i - \guide \rangle}
{ \| x_i - z \| \, \| x_i-\guide \| }
\label{eq:omega_def}
\end{equation}
% Define $v_i(z)=x_i - z$ to be the vector from $z$ to $x_i$; and similarly 
% for $v_i(\guide)$. 
% Then the angular consistency $\Omega(z,\guide)$ 
% between $z$ and $\guide$ with respect to the reference set is defined as:
% \begin{equation}
% \Omega(z,\guide) = \frac{1}{k}\sum^k_i{\frac{\langle v_i(z), v_i(\guide) \rangle}{ \|v_i(z)\| \|v_i(\guide)\| }} \label{eq:omega_def}
% \end{equation}
Fig.\ \ref{sf:om_fc7_train} and \ref{sf:om_pool5_train} show histograms 
of $\Omega(\adv,\guide)$ compared to 
$\Omega(n_c,\guide)$ where $n_c \in N_c$ and
$\Omega(n_f,\guide)$ where $n_f \in N_f$.
Note that maximum angular consistency is $1$, in which case the point 
behaves like $\guide$. Other than differences in scaling and upper 
bound, the angular consistency plots \ref{sf:om_fc7_train} and 
\ref{sf:om_pool5_train} are strikingly similar to those for the likelihood 
comparisons in the first two columns of Fig.\ \ref{fig:local_inlier}, 
supporting the conclusion that $\adv$ is an inlier with respect to 
representations of natural images.

%% \yanshuai{Add description about this measure: here, basically we are using the 15 random points of $N_{20}$ as reference landmarks, and draw rays from other points to them, and look at how corresponding rays differ for $g$ and another point by measuring cosine distance. An angular consistency value of 1 means the two points are probably the same.}

\begin{figure}[t]
\centering
\begin{subfigure}[t]{.26\linewidth}
\includegraphics[width=\linewidth]{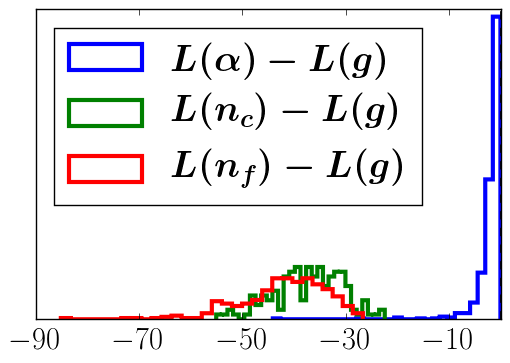}
\caption{$\Delta L$, FC7, $\guide \in$ training} \label{sf:dl_fc7_train}
\end{subfigure}
\begin{subfigure}[t]{.26\linewidth}
\includegraphics[width=\linewidth]{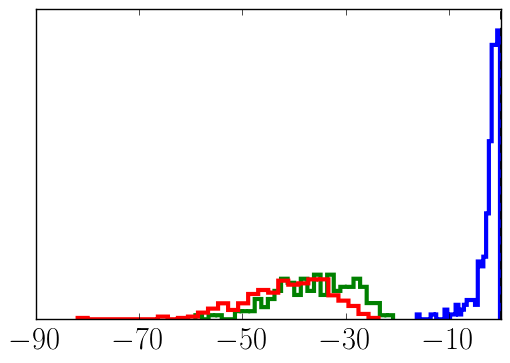}
\caption{$\Delta L$, FC7, $\guide \in$ validation} \label{sf:dl_fc7_val}
\end{subfigure}
\hspace*{0.5cm}
\begin{subfigure}[t]{.26\linewidth}
\includegraphics[width=\linewidth]{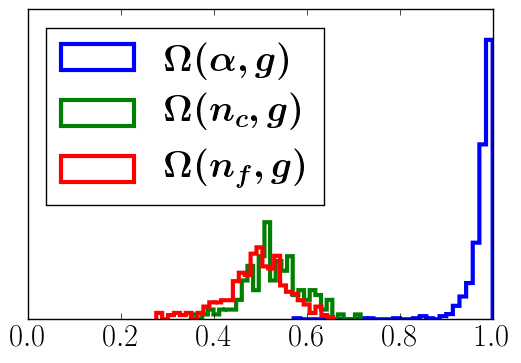}
\caption{$\Omega$, FC7, $\guide \in$ training} \label{sf:om_fc7_train}
\end{subfigure}

\begin{subfigure}[t]{.26\linewidth}
\includegraphics[width=\linewidth]{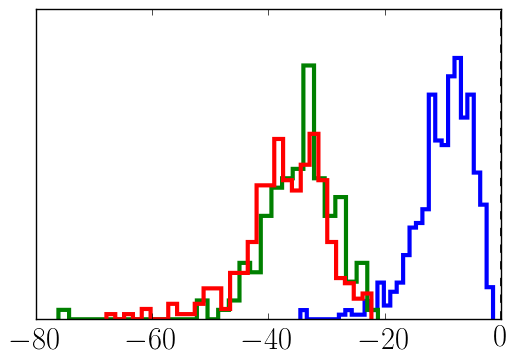}
\caption{$\Delta L$, P5, $\guide \in$ training} \label{sf:dl_pool5_train}
\end{subfigure}
\begin{subfigure}[t]{.26\linewidth}
\includegraphics[width=\linewidth]{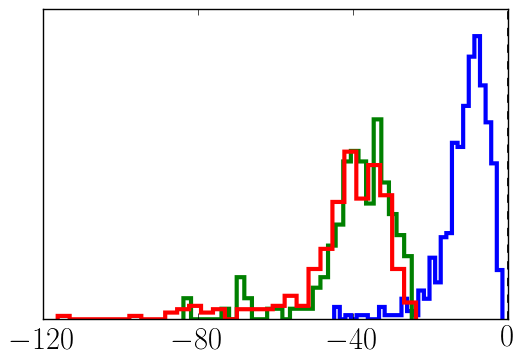}
\caption{$\Delta L$, P5, $\guide \in$ validation} \label{sf:dl_pool5_val}
\end{subfigure}
\hspace*{0.5cm}
\begin{subfigure}[t]{.26\linewidth}
\includegraphics[width=\linewidth]{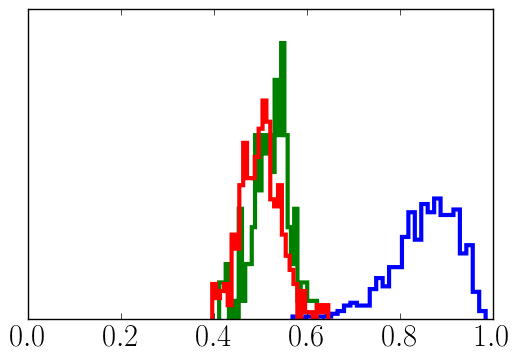}
\caption{$\Omega$, P5, $\guide \in$ training} \label{sf:om_pool5_train}
\end{subfigure}

\vspace*{-0.1cm}
\caption{Manifold inlier analysis: the first two columns 
(\ref{sf:dl_fc7_train},\ref{sf:dl_fc7_val},\ref{sf:dl_pool5_train},\ref{sf:dl_pool5_val}) 
for results of manifold tangent space analysis, showing distribution of 
difference in log likelihood of a point and $\guide$, $\Delta 
L(\cdot,\guide)=L(\cdot)-L(\guide)$; the last column 
(\ref{sf:om_fc7_train}),(\ref{sf:om_pool5_train}) for angular consistency 
analysis, showing distribution of angular consistency $\Omega(\cdot,g)$, 
between a point and $\guide$. See Eqn.\ \ref{eq:omega_def} for definitions.} 
\label{fig:local_inlier}
\vspace*{-0.2cm}
\end{figure}

%% \comment{comparing the log likelihood ($L$) of various points under a generative model locally built around the guide image feature. (Adv, \textcolor{blue}{BLUE}) $L(a)-L(g)$, (NNc, \textcolor{green}{GREEN}) $L(n_c)-L(g)$, (NNf, \textcolor{red}{RED}) $L(n_f)-L(g)$; where $g$ is a guide image feature; $a$ is an adversarial image feature; $n_c$ a random top-$20$ near neighbor of $g$ that is not used in building the local PPCA; $n_f$ a near neighbor of $g$ in the rank $21$ to $30$.}}

\subsection{Comparisons and analysis}
\label{sec:comparison}
\vspace*{-0.2cm}

We now compare our feature adversaries to images created to optimize 
mis-classification \citep{SzegedyElatICLR2014}, in part to illustrate 
qualitative differences.
We also investigate if the linearity hypothesis for mis-classification 
adversaries of \cite{GoodfellowEtalICLR2015} is consistent with and 
explains  with our class of adversarial examples.
We hereby refer to our results as {\em feature adversaries via 
optimization (feature-opt)}. The adversarial images designed to trigger 
mis-classification via optimization \citep{SzegedyElatICLR2014}, 
described briefly in Sec.~\ref{related}, are referred to as 
{\em label adversaries via optimization (label-opt)}.

\vspace*{-0.25cm}
\paragraph{Comparison to label-opt:} 

\setcounter{subfigure}{0}
\begin{figure*}[t!]
%\centering
%\begin{subfigure}[t]{.5\linewidth}
\centering
\renewcommand{\arraystretch}{1}
\setlength\tabcolsep{5pt}
\setlength{\abovecaptionskip}{5pt}
%\vspace*{-2.3cm}
\begin{subfigure}[t]{.001\linewidth}
\end{subfigure}
\begin{tabular}{
>{\centering\arraybackslash}m{0.6\linewidth}
>{\centering\arraybackslash}m{0.38\linewidth}}
%\hspace*{-1cm}
%\vspace{-0.45cm}
\setlength\tabcolsep{2pt}
\begin{tabular}{
>{\centering\arraybackslash}m{0.45\linewidth}
>{\centering\arraybackslash}m{0.45\linewidth}}
\includegraphics[width=\linewidth]{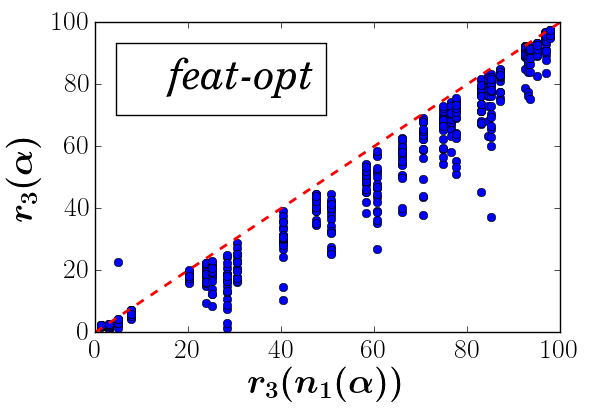} &
\includegraphics[width=\linewidth]{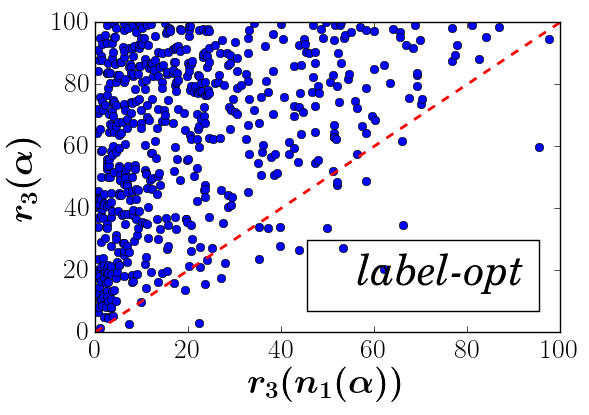} 
%\end{tabular}
\end{tabular}
%\vspace*{-0.15cm}
\captionof{subfigure}{Rank of adversaries vs rank of $n_1(\alpha)$: 
Average distance of $3$-NNs is used to rank all points in predicted 
class (excl.\ guide). Adversaries with same horizontal coordinate
share the same guide.}
\label{fig:scatter_rank}
&
%\hspace*{0.2cm}
\vspace*{-0.75cm}
\includegraphics[width=.7\linewidth]{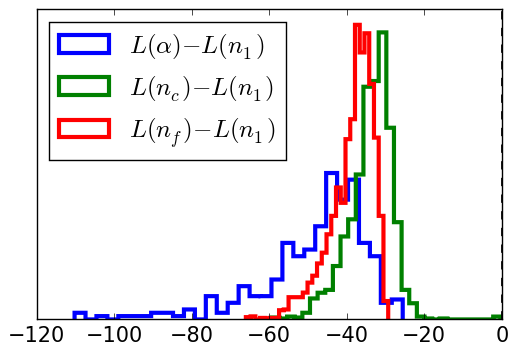} 
%\begin{subfigure}[t]{.245\linewidth}
\vspace*{0.15cm}
\captionof{subfigure}{Manifold analysis for label-opt adversaries, at layer 
FC7, with tangent plane through $n_1(\alpha)$.}
\label{fig:labeladv_PPCA_Dist}

%\end{subfigure}
%\caption{\small{label-opt, Gaussian at $n_1(\adv)$}}
%% \begin{subfigure}[t]{.325\linewidth}
%% \begin{center}
%% \includegraphics[width=0.95\linewidth]{./imgs/localpca_improved_compare_true_nn__cosine_20_90_fc7fgrad_feat_ll_hists.png}
%% \end{center}
%% \caption{\small{feat-fgrad, Gaussian at $\guide$}}
%% \end{subfigure}
%% \caption{Manifold analysis for label-opt adversaries, at layer 
%% FC7, with tangent plane through $n_1(\alpha)$}% \yanshuai{show this for label-fgrad and feat-fgrad}}

%% \begin{subfigure}[t]{.245\linewidth}
%% \includegraphics[width=\linewidth]{./imgs/rank_all_fgrad_label.png}
%% \end{subfigure}
%% \begin{subfigure}[t]{.245\linewidth}
%% \includegraphics[width=\linewidth]{./imgs/rank_all_fgrad_fc7.png}
%% \end{subfigure}
\end{tabular}
\vspace*{-0.2cm}
% To take care of the mess that tabular has made!
\addtocounter{figure}{-1}
\vspace*{-0.25cm}
\caption{Label-opt and feature-opt PPCA and rank measure comparison plots.}
\label{fig:compare}
\end{figure*}

%\yanshuai{adv on labels and show rep on the layer below is out-of-subspace;
%show this using lpca and sparsity pattern;
%}
%\yanshuai{previous TBD “Understanding adv and causes" section removed and merged here.}
%\yanshuai{Introduce fast gradient method}

To demonstrate that label-opt differs qualitatively from feature-opt, 
we report three empirical results. 
First, we rank $\adv$, $\guide$, and other points assigned the same 
class label as $\guide$, according to their average distance to three 
nearest neighbours, as in Sec.~\ref{sec:closeness}. 
Fig.~\ref{fig:scatter_rank} shows rank of $\adv$ versus rank of its 
nearest neighbor-$n_1(\adv)$ for the two types of adversaries. Unlike 
feature-opt, for label-opt, the rank of $\adv$ does not correlate well 
with the rank of $n_1(\adv)$.  In other words, for feature-opt $\adv$ is 
close to $n_1(\adv)$, while for label-opt it is not.

Second, we use the manifold PPCA approach in Sec.~\ref{sec:natural_reps}.
Comparing to peaked histogram of standardized likelihood of feature-opt 
shown in Fig.~\ref{fig:local_inlier}, Fig.~\ref{fig:labeladv_PPCA_Dist} 
shows that label-opt examples are not represented well by the Gaussian 
around the first NN of $\adv$.

%% And for the case of feat-fgrad, neither the Gaussian around the first NN nor $\guide$ explain 
%% $\adv$ well. \yanshuai{complete Fig. \ref{fig:labeladv_PPCA_Dist} with the 
%% subfigures needed for label-fgrad and feat-fgrad.}
%%And for the case of feat-fgrad, it is not well explained by the Gaussian around $\guide$.

Third, we analyze the sparsity  patterns on different DNN layers for 
different adversarial construction methods. It is well known that DNNs 
with ReLU activation units produce sparse activations 
(\cite{AISTATS2011_GlorotBB11}).  Therefore, if the degree of sparsity 
increases after the adversarial perturbation, the adversarial example 
is using additional paths to manipulate the resulting represenation. 
We also investigate how many activated units are shared between the source 
and the adversary, by computing the intersection over union {\em $I/U$} of 
active units. If the $I/U$ is high on all layers, then two represenations 
share most active paths. On the other hand, if $I/U$ is low, 
while the degree of sparsity remains the same, then the adversary must have 
closed some activation paths and opened new ones. In Table~\ref{tb:sparsity}, 
$\Delta S$ is the difference between the proportion of non-zero activations on 
selected layers between the source image represenation for the two types 
of adversaries. One can see that for all except FC$7$ of label-opt, the 
difference is significant. The column ``$I/U$ with $\source$'' also shows that 
feature-opt uses very different activation paths from $\source$ when compared 
to label-opt. 

%% In the last column ``$I/U$ with $\guide$'', feature-opt shows much 
%% higher $I/U$ with the 
%% guide representation on the targeted layer FC$7$ comparing to 
%% the one obtained by feat-fgrad.

\vspace*{-0.25cm}
\paragraph{Testing The Linearity Hypothesis for feature-opt:}
\cite{GoodfellowEtalICLR2015} suggests that the existence of label adversaries
is a consequence of networks being too linear.  If this linearity hypothesis 
applies to our class of adversaries, it should be possible to linearize the 
DNN around the source image, and then obtain similar adversaries via 
optimization. Formally, let $J_s = J(\phi(I_s))$ be the Jacobian matrix 
of the internal layer encoding with respect to source image input.  
Then, the 
linearity hypothesis implies $\phi(I) \approx \phi(I_s) + J_s^{\T}(I-I_s)$.
Hence, we optimize $\| \phi(I_s) + J_s^{\T}(I-I_s) - \phi(I_g)\|^2_2$ 
subject to the same infinity norm constraint in Eqn.\ \ref{infnorm_bound}.
We refer to these adversaries as {\em feature-linear}.  

As shown in Fig.~\ref{fig:delta_layer}, such adversaries do not get 
particularly close to the guide.  They get no closer than 80\%, while
for {\em feature-opt} the distance is reduced to $50\%$ or less 
for layers down to C2.  Note that unlike {\em feature-opt}, the objective 
of {\em feature-linear} does not guarantee a reduction in distance 
when the constraint on $\delta$ is relaxed.  These results suggest that 
the linearity hypothesis may not explain the existence of {\em feature-opt} 
adversaries.

\comment{
%% \begin{figure}[t]
%% \centering
%% %\begin{subfigure}[t]{.325\linewidth}
%% \begin{center}
%% \includegraphics[width=0.325\linewidth]{./imgs/label-opt_1nn_ll_hists.png}
%% \end{center}
%% \caption{\small{label-opt, Gaussian at $n_1(\adv)$}}
%% %\end{subfigure}

%% %% \begin{subfigure}[t]{.325\linewidth}
%% %% \begin{center}
%% %% \includegraphics[width=0.95\linewidth]{./imgs/label-fgrad_1nn_ll_hists.png}
%% %% \end{center}
%% %% \caption{\small{label-fgrad, Gaussian at $n_1(\adv)$}}
%% %% \end{subfigure}
%% %% \begin{subfigure}[t]{.24\linewidth}
%% %% \includegraphics[width=\linewidth]{./imgs/localpca_improved_compare_true_nn__cosine_20_90_fc7fgrad_feat_ll_hists.png}
%% %% \caption{\small{fc7, feat-fgrad, Gaussian at $n_1(\adv)$  (place-holder)}}
%% %% \end{subfigure}
%% %% \begin{subfigure}[t]{.325\linewidth}
%% %% \begin{center}
%% %% \includegraphics[width=0.95\linewidth]{./imgs/localpca_improved_compare_true_nn__cosine_20_90_fc7fgrad_feat_ll_hists.png}
%% %% \end{center}
%% %% \caption{\small{feat-fgrad, Gaussian at $\guide$}}
%% %% \end{subfigure}
%% \caption{Manifold analysis for label-opt adversaries, at layer 
%% FC7, with tangent plane through $n_1(\alpha)$}% \yanshuai{show this for label-fgrad and feat-fgrad}}
%% \label{fig:labeladv_PPCA_Dist}
%% \end{figure}
}
\begin{table*}[t]
\renewcommand{\arraystretch}{1}
\setlength\tabcolsep{2pt}
\setlength{\abovecaptionskip}{25pt}
%\vspace*{-2.3cm}
%\begin{subfigure}[t]{.001\linewidth}
%\end{subfigure}
\begin{tabular}{
>{\centering\arraybackslash}m{0.48\linewidth}
>{\centering\arraybackslash}m{0.02\linewidth}
>{\centering\arraybackslash}m{0.48\linewidth}}
\setlength\tabcolsep{2pt}
%\begin{table*}[t]
%    \resizebox{\linewidth}{!}{
%\centering
\vspace*{-0.2cm}
\begin{footnotesize}
\begin{tabular}{l|cc|cc|}
\cline{2-5}
   & \multicolumn{2}{c|}{$\Delta S$}   & \multicolumn{2}{c|}{I/U with s}  \\ \cline{2-5}
                            & \multicolumn{1}{c|}{feature-opt} & \multicolumn{1}{c|}{label-opt}  & \multicolumn{1}{c|}{feature-opt} & \multicolumn{1}{c|}{label-opt} \\ \hline
\multicolumn{1}{|l|}{FC7}  & $7 \pm 7$ & $13 \pm 5$ & $\boldsymbol{12 \pm 4}$ & $39 \pm 9$  \\ \cline{1-1}
\multicolumn{1}{|l|}{C5} & $0 \pm 1$ & $0 \pm 0$ &  $33 \pm 2$ & $70 \pm 5$   \\ \cline{1-1}
\multicolumn{1}{|l|}{C3} & $2 \pm 1$ & $0 \pm 0$ &   $60 \pm 1$ & $85 \pm 3$  \\ \cline{1-1}
\multicolumn{1}{|l|}{C1} &  $0 \pm 0$ & $0 \pm 0$ &   $78 \pm 0$ & $94 \pm 1$   \\ \hline
\end{tabular}
\end{footnotesize}
\vspace*{-0.5cm}
\captionof{table}{Sparsity analysis: Sparsity is quantified 
as a percentage of the size of each layer.}
\label{tb:sparsity}& &
\setlength{\abovecaptionskip}{3pt}
%}
%\caption{Sparsity pattern analysis: all numbers in percentage.}
%\label{tb:sparsity} &
%\end{table*} &
%\vspace*{-.6cm}

\vspace*{-.2cm}
\includegraphics[width=.625\linewidth]{./imgs/delta_layers.png}
\captionof{figure}
{Distance ratio $\left. d(\adv,\!\guide) \middle/
d(\source,\!\guide)\right.$ vs $\delta$.
C$2$, C$3$, P$5$, F$7$ are for {\it feature-opt} adversaries;
$\ell$-f7 denotes FC7 distances for {\it feature-linear}.}
\label{fig:delta_layer}
\end{tabular}
\vspace*{-0.5cm}
\end{table*}

\vspace*{-0.25cm}
\paragraph{Networks with Random Weights:}
We further explored whether the existence of {\em feature-opt} adversaries 
is due to the learning algorithm and the training set, or to the structure 
of deep networks per se. For this purpose, we randomly initialized layers of 
Caffenet with orthonormal weights.  We then optimized for adversarial images 
as above, and looked at distance ratios (as in Fig.\ \ref{fig:H_Dist}).  
Interestingly, the distance ratios for FC$7$ and Norm$2$ are similar to 
Fig.~\ref{fig:delta_layer} with at most $2\%$ deviation.  On C$2$, the results 
are at most $10\%$ greater than those on C$2$ for the trained Caffenet.  
We note that both Norm$2$ and C$2$ are overcomplete representations 
of the input. The table of distance ratios can be found in the 
Supplementary Material.
These results with random networks suggest that the existence of 
{\em feature-opt} adversaries may be a property of the network architecture.

\comment{This observation shows that the model itself is vulnerable to {\em 
feature-opt} adversaries, rather than just a specific trained network.}

% , whereas 
% our main experiments with trained networks showed various properties 
% of these adversaries that have practical consequences for computer vision. 

%\input{exp_understanding}

\section{Discussion}
\vspace*{-0.1cm}

We introduce a new method for generating adversarial images that appear 
perceptually similar to a given source image, but whose deep representations 
mimic the characteristics of natural guide images. 
Indeed, the adversarial images have representations at intermediate 
layers appear quite natural and very much like the guide images used 
in their construction.
We demonstrate empirically that these imposters capture the generic 
nature of their guides at different levels of deep representations.  This 
includes their proximity to the guide, and their locations in high density 
regions of the feature space.
We show further that such properties are not shared by other
categories of adversarial images.

\comment{{{
The approach for generating adversarial images with internal representations 
that mimic those of other images has worked well with a remarkably broad 
class of images, including images from training and test sets, and images 
of all classes in the Imagenet and Places datasets. 
% Understanding the root causes of this phenomena remains unanswered.
%since the network is not particularly deep, and the data are not typical of

%% One possible cause may be that the fine-tuned net only exploits a subspace
%% of the FC$7$ representation, so during fine-tuning there may be distortions
%% to features outside that subspace that provide marginal gains in classification.
%% As a consequence, Euclidean distance in FC$7$ will no longer be a useful loss 
%% function. That is, the fine-tuned representations are no longer generic 
%% representations for natural images since features are projected out 
%% before the final softmax layer. 
%% There is some evidence in favour of this view, as we find that average distance of three NNs of the $\adv$ is always one of the smallest comparing to other points in the class, meaning that optimization for $\adv$ is somehow stuck in a high density region in the space. This could be due to the distortion of the representation space by fine-tuning. 
%% \david{I am not sure this discussion of Flicker Style is
%% clear enough to justify some much space in the paper?} 

%These failures suggest that the adversarial phenomena reported here depend both
%on having deep networks and a broad class of natural image inputs. Although receptive field
%\fartash{This needs support from the new experiments we did based on Ian 
%Goodfellow’s comments}
}}}

We also find that the linearity hypothesis \citep{GoodfellowEtalICLR2015} 
does not provide an obvious explanation for these new adversarial phenomena.
It appears that the existence of these adversarial images is not predicated 
on a network trained with natural images per se.  For example, results
on random networks indicate that the structure of the network itself may 
be one significant factor.  Nevertheless, further experiments and analysis 
are required to determine the true underlying reasons for this discrepancy 
between human and DNN representations of images.

Another future direction concerns the exploration of failure cases we 
observed in optimizing feature adversaries. As mentioned in supplementary 
material, such cases involve images of hand-written digits, and networks 
that are fine-tuned with images from a narrow domain (e.g., the Flicker 
Style dataset). Such failures suggest that our adversarial phenomena may 
be due to factors such as network  depth, receptive field size, or the 
class of natural images used.  Since our aim here was to analyze the 
representation of well-known networks, we leave the exploration of these
factors to future work.
Another interesting question concerns whether existing discriminative 
models might be trained to detect feature adversaries.  Since training 
such models requires a diverse and relatively large dataset of adversarial 
images we also leave this to future work.

\begin{footnotesize}
\paragraph{ACKNOWLEDGMENTS}  Financial support for this research was 
provided, in part, by MITACS, NSERC Canada, and the Canadian Institute 
for Advanced Research (CIFAR). We would like to thank Foteini Agrafioti 
for her support. We would also like to thank Ian Goodfellow, Xavier Boix, 
as well as the anoynomous reviewers for helpful feedback.
\end{footnotesize}

\bibliography{adv}

\begin{thebibliography}{15}
\providecommand{\natexlab}[1]{#1}
\providecommand{\url}[1]{\texttt{#1}}
\expandafter\ifx\csname urlstyle\endcsname\relax
  \providecommand{\doi}[1]{doi: #1}\else
  \providecommand{\doi}{doi: \begingroup \urlstyle{rm}\Url}\fi

\bibitem[Chatfield et~al.(2014)Chatfield, Simonyan, Vedaldi, and
  Zisserman]{chatfield2014return}
Chatfield, K., Simonyan, K., Vedaldi, A., and Zisserman, A.
\newblock Return of the devil in the details: Delving deep into convolutional
  nets.
\newblock In \emph{BMVC}, 2014.

\bibitem[Deng et~al.(2009)Deng, Dong, Socher, Li, Li, and
  Fei-Fei]{deng2009imagenet}
Deng, J, Dong, W, Socher, R, Li, LJ, Li, K, and Fei-Fei, L.
\newblock Imagenet: A large-scale hierarchical image database.
\newblock In \emph{IEEE CVPR}, pp.\  248--255, 2009.

\bibitem[Fawzi et~al.(2015)Fawzi, Fawzi, and Frossard]{FawziEtalICLR2015}
Fawzi, A, Fawzi, O, and Frossard, P.
\newblock Fundamental limits on adversarial robustness.
\newblock In \emph{ICML}, 2015.

\bibitem[Glorot et~al.(2011)Glorot, Bordes, and Bengio]{AISTATS2011_GlorotBB11}
Glorot, X, Bordes, A, and Bengio, Y.
\newblock Deep sparse rectifier neural networks.
\newblock In \emph{AISTATS}, volume~15, pp.\  315--323, 2011.

\bibitem[Goodfellow et~al.(2014)Goodfellow, Shlens, and
  Szegedy]{GoodfellowEtalICLR2015}
Goodfellow, IJ, Shlens, J, and Szegedy, C.
\newblock Explaining and harnessing adversarial examples.
\newblock In \emph{ICLR (arXiv:1412.6572)}, 2014.

\bibitem[Gu \& Rigazio(2014)Gu and Rigazio]{GuRigazioNIPSWorkshop2014}
Gu, S and Rigazio, L.
\newblock Towards deep neural network architectures robust to adversarial
  examples.
\newblock In \emph{Deep Learning and Representation Learning Workshop
  (arXiv:1412.5068)}, 2014.

\bibitem[Jia et~al.(2014)Jia, Shelhamer, Donahue, Karayev, Long, Girshick,
  Guadarrama, and Darrell]{jia2014caffe}
Jia, Y, Shelhamer, E, Donahue, J, Karayev, S, Long, J, Girshick, R, Guadarrama,
  S, and Darrell, T.
\newblock Caffe: Convolutional architecture for fast feature embedding.
\newblock In \emph{ACM Int.\ Conf.\ Multimedia}, pp.\  675--678, 2014.

\bibitem[Krizhevsky et~al.(2012)Krizhevsky, Sutskever, and
  Hinton]{krizhevsky2012imagenet}
Krizhevsky, A, Sutskever, I, and Hinton, GE.
\newblock Imagenet classification with deep convolutional neural networks.
\newblock In \emph{NIPS}, pp.\  1097--1105, 2012.

\bibitem[Mahendran \& Vedaldi(2014)Mahendran and
  Vedaldi]{MahendranVedaldiCVPR2015}
Mahendran, A and Vedaldi, A.
\newblock Understanding deep image representations by inverting them.
\newblock In \emph{IEEE CVPR (arXiv:1412.0035)}, 2014.

\bibitem[Nguyen et~al.(2015)Nguyen, Yosinski, and Clune]{NguyenEtAlCVPR2015}
Nguyen, A, Yosinski, J, and Clune, J.
\newblock Deep neural networks are easily fooled: High confidence predictions
  for unrecognizable images.
\newblock In \emph{IEEE CVPR (arXiv:1412.1897)}, 2015.

\bibitem[Szegedy et~al.(2014)Szegedy, Zaremba, Sutskever, Bruna, Erhan,
  Goodfellow, and Fergus]{SzegedyElatICLR2014}
Szegedy, C, Zaremba, W, Sutskever, I, Bruna, J, Erhan, D, Goodfellow, I, and
  Fergus, R.
\newblock Intriguing properties of neural networks.
\newblock In \emph{ICLR (arXiv:1312.6199)}, 2014.

\bibitem[Szegedy et~al.(2015)Szegedy, Liu, Jia, Sermanet, Reed, Anguelov,
  Erhan, Vanhoucke, and Rabinovich]{szegedy2014going}
Szegedy, C, Liu, W, Jia, Y, Sermanet, P, Reed, S, Anguelov, D, Erhan, D,
  Vanhoucke, V, and Rabinovich, A.
\newblock Going deeper with convolutions.
\newblock In \emph{CVPR}, 2015.

\bibitem[Tabacof \& Valle(2015)Tabacof and Valle]{Tabacof2015exploring}
Tabacof, P and Valle, E.
\newblock Exploring the space of adversarial images.
\newblock \emph{arXiv preprint arXiv:1510.05328}, 2015.

\bibitem[Wang et~al.(2004)Wang, Bovik, Sheikh, and
  Simoncelli]{WangEtalPAMI2004}
Wang, Z, Bovik, AC, Sheikh, HR, and Simoncelli, EP.
\newblock Image quality assessment: From error visibility to structural
  similarity.
\newblock \emph{IEEE Trans.\ PAMI}, 3\penalty0 (4):\penalty0 600--612, 2004.

\bibitem[Zhou et~al.(2014)Zhou, Lapedriza, Xiao, Torralba, and
  Oliva]{zhou2014learning}
Zhou, B, Lapedriza, A, Xiao, J, Torralba, A, and Oliva, A.
\newblock Learning deep features for scene recognition using places database.
\newblock In \emph{NIPS}, pp.\  487--495, 2014.

\end{thebibliography}
\bibliographystyle{iclr2016_conference}
\vfill

\newpage
\beginsupplement
\section*{Supplementary Material}
\label{}

\subsection{Illustration of the idea}
\begin{figure}[h]
\centering
\includegraphics[width=\linewidth]{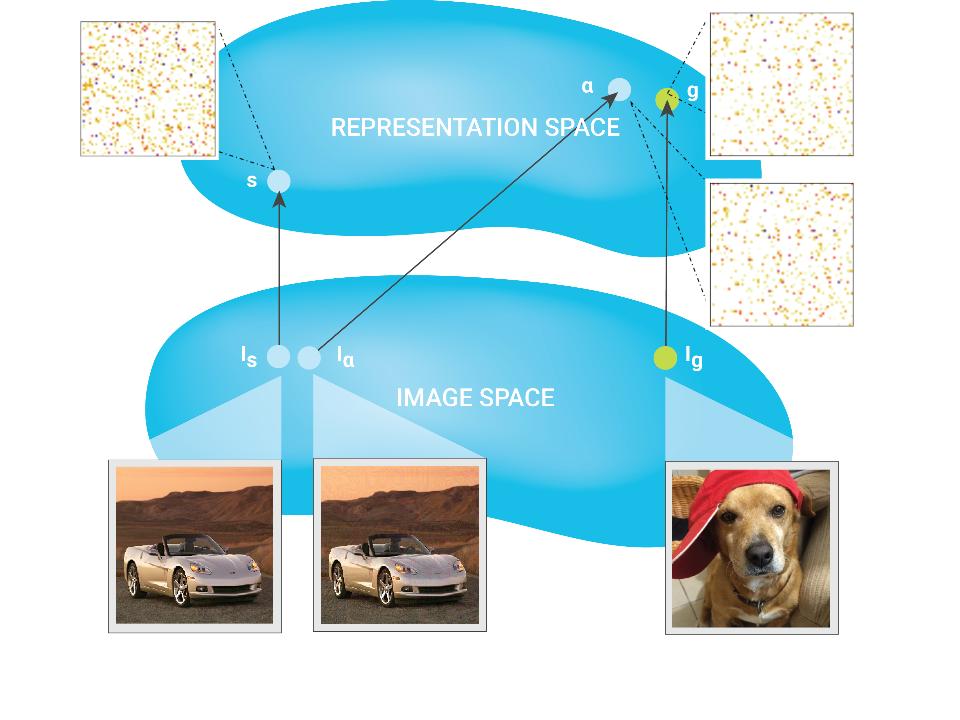}
\caption{Summary of the main idea behind the paper.} \label{fig:illustrate}
\vspace*{-0.2cm}
\end{figure}

Fig.~\ref{fig:illustrate} illustrates the achieved goal in this paper. The 
image of the fancy car on the left is a training example from the ILSVRC 
dataset. On the right of it, there is an adversarial image that was generated 
by guiding the source image by an image of Max (the dog). While the two fancy 
car images are very close in image space, the activation pattern of the 
adversarial car is almost identical to that of Max.  This shows that the 
mapping from the image space to the representation space is such that for each 
natural image, there exists a point in a small neighborhood in the image space 
that is mapped by the network to a point in the representation space that is in 
a small neighborhood of the representation of a very different natural image.

\subsection{Datasets for Empirical Analysis}

Unless stated otherwise,  we have used the following two sets of source and 
guide images. The first set is used for experiments on layer FC$7$ and the 
second set is used for computational expedience on other layers (e.g.  P$5$). 
The source images are guided by all guide images to show that the convergence 
does not depend on the class of images. To simplify the reporting of 
classification behavior, we only used guides from training set whose labels are 
correctly predicted by Caffenet.

In both sets we used $20$ source images, with five drawn at random from each of 
the ILSVRC train, test and validation sets, and five more selected manually 
from Wikipedia and the ILSVRC validation set to provide greater diversity.  The 
guide set for the first set consisted of three images from each of $1000$
classes, drawn at random from ILSVRC training images, and another $30$ images 
from each of the validation and test sets. For the second set, we drew guide 
images from just $100$ classes.

\subsection{Examples of Adversaries}
\begin{figure*}[h!]
\centering
\renewcommand{\arraystretch}{1}
\setlength\tabcolsep{2pt}
\begin{tabular}{|
>{\centering\arraybackslash}m{0.15\linewidth}
>{\centering\arraybackslash}m{0.15\linewidth}
>{\centering\arraybackslash}m{0.15\linewidth}
>{\centering\arraybackslash}m{0.005\linewidth} |
>{\centering\arraybackslash}m{0.005\linewidth}
>{\centering\arraybackslash}m{0.15\linewidth}
>{\centering\arraybackslash}m{0.15\linewidth}
>{\centering\arraybackslash}m{0.15\linewidth}|}
\hline
{\footnotesize Source} &
{\footnotesize $I_{\adv}^{\text{FC}7}\!,\delta\!=\!5$}  &{\footnotesize Guide} 
& & 
& {\footnotesize Source} &
{\footnotesize $I_{\adv}^{P5}\!,\delta\!=\!10$}  & {\footnotesize Guide} 
\\[1ex] \hline

\includegraphics[width=\linewidth, height=.75\linewidth]{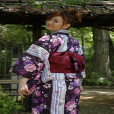} &
\includegraphics[width=\linewidth, height=.75\linewidth]{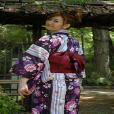} &
\includegraphics[width=\linewidth, height=.75\linewidth]{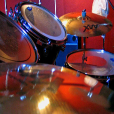}  & 
& & 
\includegraphics[width=\linewidth, height=.75\linewidth]{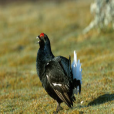} &
\includegraphics[width=\linewidth, height=.75\linewidth]{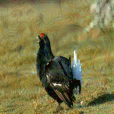} &
\includegraphics[width=\linewidth, height=.75\linewidth]{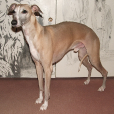}  \\
\includegraphics[width=\linewidth, height=.75\linewidth]{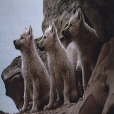} &
\includegraphics[width=\linewidth, height=.75\linewidth]{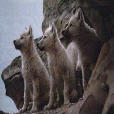} &
\includegraphics[width=\linewidth, height=.75\linewidth]{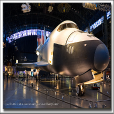}  &
& & 
\includegraphics[width=\linewidth, height=.75\linewidth]{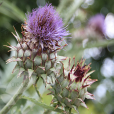} &
\includegraphics[width=\linewidth, height=.75\linewidth]{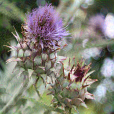} &
\includegraphics[width=\linewidth, height=.75\linewidth]{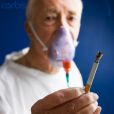}  \\
\includegraphics[width=\linewidth, height=.75\linewidth]{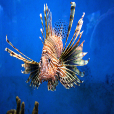} &
\includegraphics[width=\linewidth, height=.75\linewidth]{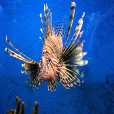} &
\includegraphics[width=\linewidth, height=.75\linewidth]{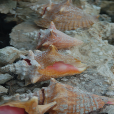}  &
 & &
\includegraphics[width=\linewidth, height=.75\linewidth]{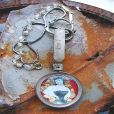} &
\includegraphics[width=\linewidth, height=.75\linewidth]{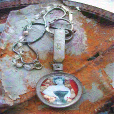} &
\includegraphics[width=\linewidth, height=.75\linewidth]{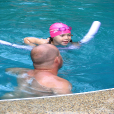}  \\
\includegraphics[width=\linewidth, height=.75\linewidth]{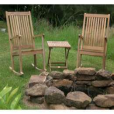} &
\includegraphics[width=\linewidth, height=.75\linewidth]{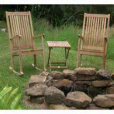} &
\includegraphics[width=\linewidth, height=.75\linewidth]{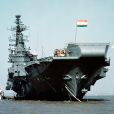}  &
& &
\includegraphics[width=\linewidth, height=.75\linewidth]{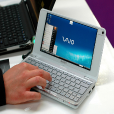} &
\includegraphics[width=\linewidth, height=.75\linewidth]{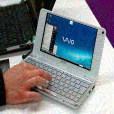} &
\includegraphics[width=\linewidth, height=.75\linewidth]{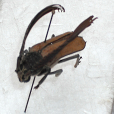}  \\
\includegraphics[width=\linewidth, height=.75\linewidth]{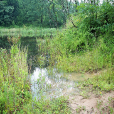} &
\includegraphics[width=\linewidth, height=.75\linewidth]{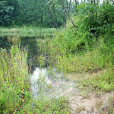} &
\includegraphics[width=\linewidth, height=.75\linewidth]{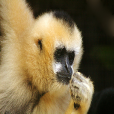}  &
& &
\includegraphics[width=\linewidth, height=.75\linewidth]{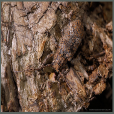} &
\includegraphics[width=\linewidth, height=.75\linewidth]{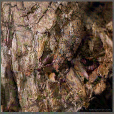} &
\includegraphics[width=\linewidth, height=.75\linewidth]{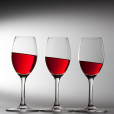}  \\
\includegraphics[width=\linewidth, height=.75\linewidth]{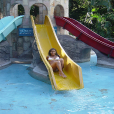} &
\includegraphics[width=\linewidth, height=.75\linewidth]{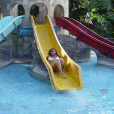} &
\includegraphics[width=\linewidth, height=.75\linewidth]{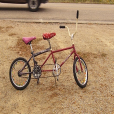}  &
& &
\includegraphics[width=\linewidth, height=.75\linewidth]{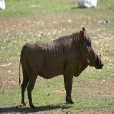} &
\includegraphics[width=\linewidth, height=.75\linewidth]{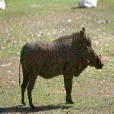} &
\includegraphics[width=\linewidth, height=.75\linewidth]{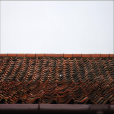}  \\
\includegraphics[width=\linewidth, height=.75\linewidth]{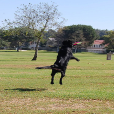} &
\includegraphics[width=\linewidth, height=.75\linewidth]{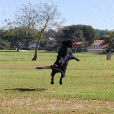} &
\includegraphics[width=\linewidth, height=.75\linewidth]{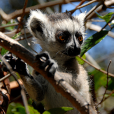}  &
& &
\includegraphics[width=\linewidth, height=.75\linewidth]{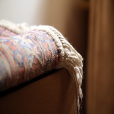} &
\includegraphics[width=\linewidth, height=.75\linewidth]{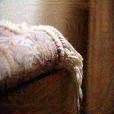} &
\includegraphics[width=\linewidth, height=.75\linewidth]{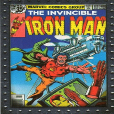}  \\
\includegraphics[width=\linewidth, height=.75\linewidth]{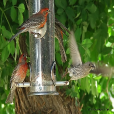} &
\includegraphics[width=\linewidth, height=.75\linewidth]{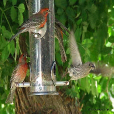} &
\includegraphics[width=\linewidth, height=.75\linewidth]{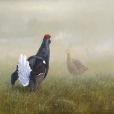}  &
& &
\includegraphics[width=\linewidth, height=.75\linewidth]{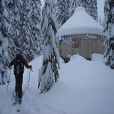} &
\includegraphics[width=\linewidth, height=.75\linewidth]{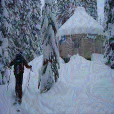} &
\includegraphics[width=\linewidth, height=.75\linewidth]{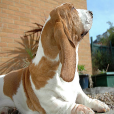}  \\
\includegraphics[width=\linewidth, height=.75\linewidth]{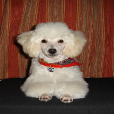} &
\includegraphics[width=\linewidth, height=.75\linewidth]{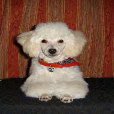} &
\includegraphics[width=\linewidth, height=.75\linewidth]{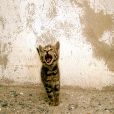}  &
& &
\includegraphics[width=\linewidth, height=.75\linewidth]{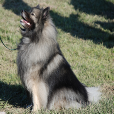} &
\includegraphics[width=\linewidth, height=.75\linewidth]{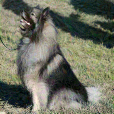} &
\includegraphics[width=\linewidth, height=.75\linewidth]{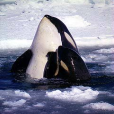}  \\
\includegraphics[width=\linewidth, height=.75\linewidth]{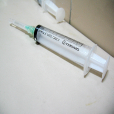} &
\includegraphics[width=\linewidth, height=.75\linewidth]{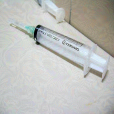} &
\includegraphics[width=\linewidth, height=.75\linewidth]{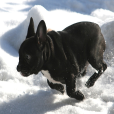}  &
& &
\includegraphics[width=\linewidth, height=.75\linewidth]{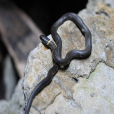} &
\includegraphics[width=\linewidth, height=.75\linewidth]{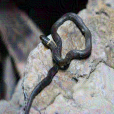} &
\includegraphics[width=\linewidth, height=.75\linewidth]{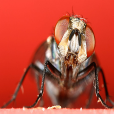}  \\
\includegraphics[width=\linewidth, height=.75\linewidth]{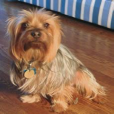} &
\includegraphics[width=\linewidth, height=.75\linewidth]{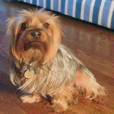} &
\includegraphics[width=\linewidth, height=.75\linewidth]{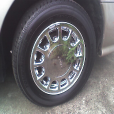}  &
& &
\includegraphics[width=\linewidth, height=.75\linewidth]{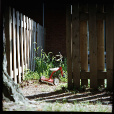} &
\includegraphics[width=\linewidth, height=.75\linewidth]{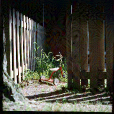} &
\includegraphics[width=\linewidth, height=.75\linewidth]{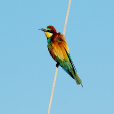}  \\
\includegraphics[width=\linewidth, height=.75\linewidth]{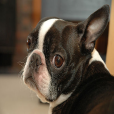} &
\includegraphics[width=\linewidth, height=.75\linewidth]{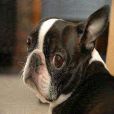} &
\includegraphics[width=\linewidth, height=.75\linewidth]{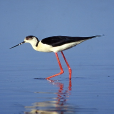}  &
& &
\includegraphics[width=\linewidth, height=.75\linewidth]{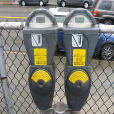} &
\includegraphics[width=\linewidth, height=.75\linewidth]{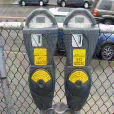} &
\includegraphics[width=\linewidth, height=.75\linewidth]{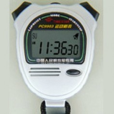}  \\
\hline
\end{tabular}
\caption{Each row shows examples of adversarial images, optimized
using different layers of Caffenet (FC$7$, P$5$), and different 
values of $\delta=(5, 10)$.  }
\label{fig:adv_caffenet_page}
\vspace*{-0.1cm}
\end{figure*}

Fig.~\ref{fig:adv_caffenet_page} shows a random sample of source and guide 
pairs along with their FC$7$ or Pool$5$ adversarial images. In none of the images the guide is perceptable in the adversary, regardless of the 
choice of source, guide or layer. The only parameter that affects the 
visibility of the noise is $\delta$.

\subsection{Dimensionality of Representations}
The main focus of this study is on the well-known Caffenet model. The layer 
names of this model and their representation dimensionalities are provided in 
Tab.~\ref{tab:caffenet}.

\begin{table}[h!]
\resizebox{\linewidth}{!}{
\centering
\begin{tabular}{|l|c|c|c|c|c|c|}
    \hline
    Layer Name &Input & Conv$2$ & Norm$2$ & Conv$3$ & Pool$5$  & FC$7$ \\ \hline
    Dimensions & $3\times 227\times 227$  & $256\times 27\times 27$ & $256\times 13\times 13$ 
    & $384\times 13\times 13$ &  $256\times 6\times 6$ & $4096$ \\ \hline
    Total & $154587$ & $186624$ & $43264$ & $64896$ & $9216$ & $4096$ \\ \hline
 \end{tabular}
}
\caption{Caffenet layer dimensions.}
\label{tab:caffenet}
\end{table}

\subsection{Results for Networks with Random Weights}
As described in Sec.~\ref{sec:comparison}, we attempt at analyzing the 
architecture of Caffenet independent of the training by initializing the model 
with random weights and generating feature adversaries. Results in 
Tab.~\ref{tab:distance} show that we can generate feature adversaries on random 
networks as well.  We use the ratio of distances of the adversary to the guide 
over the source to the guide for this analysis. In each cell, the mean and 
standard deviation of this ratio is shown for each of the three random, 
orthonormal random and trained Caffenet networks.  The weights of the random 
network are drawn from the same distribution that Caffenet is initialized with.  
Orthorgonal random weights are obtained using singular value decomposition of the regular random weights.

Results in Tab.~\ref{tab:distance} indicate that convergence on Norm$2$ and 
Conv$2$ is almost similar while the dimensionality of Norm$2$ is quite smaller 
than Conv$2$. On the other hand, Fig.~\ref{fig:delta_layer} shows that although 
Norm$2$ has smaller dimensionality than Conv$3$,  the optimization converges to 
a closer point on Conv$3$ rather than Conv$2$ and hence Norm$2$.  This means 
that the relation between dimensionality and the achieved distance of the 
adversary is not straightforward.

\begin{table*}[h!]
\resizebox{\linewidth}{!}{
\centering
\begin{tabular}{|c|c|c|c|c|c|}
\hline
Layer & $\delta=5$ & $\delta=10$ & $\delta=15$ & $\delta=20$ & $\delta=25$ \\
\hline
conv2 &
\begin{tabular}[x]{@{}c@{}} T:$0.79 \pm 0.04$ \\OR:$0.89 \pm 0.03$\\ R:$0.90 \pm 0.02$\end{tabular}
&
\begin{tabular}[x]{@{}c@{}} T:$0.66 \pm 0.06$ \\OR:$0.78 \pm 0.05$\\ R:$0.81 \pm 0.04$\end{tabular}
&
\begin{tabular}[x]{@{}c@{}} T:$0.57 \pm 0.06$ \\OR:$0.71 \pm 0.07$\\ R:$0.74 \pm 0.06$\end{tabular}
&
\begin{tabular}[x]{@{}c@{}} T:$0.50 \pm 0.07$ \\OR:$0.64 \pm 0.09$\\ R:$0.67 \pm 0.08$\end{tabular}
&
\begin{tabular}[x]{@{}c@{}} T:$0.45 \pm 0.07$ \\OR:$0.58 \pm 0.10$\\ R:$0.61 \pm 0.09$\end{tabular}
\\
\hline
norm2 &
\begin{tabular}[x]{@{}c@{}} T:$0.80 \pm 0.04$ \\OR:$0.82 \pm 0.05$\\ R:$0.85 \pm 0.03$\end{tabular}
&
\begin{tabular}[x]{@{}c@{}} T:$0.66 \pm 0.05$ \\OR:$0.69 \pm 0.08$\\ R:$0.73 \pm 0.06$\end{tabular}
&
\begin{tabular}[x]{@{}c@{}} T:$0.57 \pm 0.06$ \\OR:$0.59 \pm 0.10$\\ R:$0.63 \pm 0.08$\end{tabular}
&
\begin{tabular}[x]{@{}c@{}} T:$0.50 \pm 0.06$ \\OR:$0.51 \pm 0.11$\\ R:$0.55 \pm 0.09$\end{tabular}
&
\begin{tabular}[x]{@{}c@{}} T:$0.45 \pm 0.06$ \\OR:$0.44 \pm 0.11$\\ R:$0.48 \pm 0.10$\end{tabular}
\\
\hline
fc7 &
\begin{tabular}[x]{@{}c@{}} T:$0.32 \pm 0.10$ \\OR:$0.34 \pm 0.12$\\ R:$0.52 \pm 0.09$\end{tabular}
&
\begin{tabular}[x]{@{}c@{}} T:$0.12 \pm 0.06$ \\OR:$0.12 \pm 0.09$\\ R:$0.26 \pm 0.11$\end{tabular}
&
\begin{tabular}[x]{@{}c@{}} T:$0.07 \pm 0.04$ \\OR:$0.07 \pm 0.06$\\ R:$0.13 \pm 0.10$\end{tabular}
&
\begin{tabular}[x]{@{}c@{}} T:$0.06 \pm 0.03$ \\OR:$0.05 \pm 0.04$\\ R:$0.07 \pm 0.08$\end{tabular}
&
\begin{tabular}[x]{@{}c@{}} T:$0.05 \pm 0.02$ \\OR:$0.05 \pm 0.02$\\ R:$0.04 \pm 0.06$\end{tabular}
\\
\hline
\end{tabular}
}
\caption{Ratio of  $\left. d(\adv,\!\guide) \middle/
d(\source,\!\guide)\right.$ as $\delta$ changes from $5$ to $25$ on randomly weighted(R), orthogonal randomly weighted(OR) and trained(T) Caffenet optimized on layers Conv2, Norm2 and FC7.} 
\label{tab:distance}
\end{table*}

\subsection{Adversaries by Fast Gradient}
As we discussed in Sec.~\ref{sec:comparison}, \cite{GoodfellowEtalICLR2015} 
also proposed a method to construct label adversaries efficiently by taking 
a small step consistent with the gradient.  While this {\em fast gradient} 
method shines light on the label adversary misclassifications, and is useful 
for adversarial training, it is not relevant to whether the linearity 
hypothesis explains the feature adversaries.  Therefore we omitted the 
comparison in Sec.~\ref{sec:comparison} to fast gradient method, and continue 
the discussion here.

%\vspace*{1.0in}
The fast gradient method constructs adversaries (\cite{GoodfellowEtalICLR2015}) 
by taking the perturbation defined by $\delta \text{sign}(\nabla_I loss(f(I), 
\ell ))$, where $f$ is the classifier, and $\ell$ is an erroneous label for 
input image $I$.  We refer to the resulting adversarial examples {\em 
label-fgrad}. We can also apply the fast gradient method to an internal 
representation, i.e.\ taking the perturbation defined by $\delta 
\text{sign}(\nabla_I \| \phi(I) - \phi(I_g) \|^2)$. We call this
type {\em feature adversaries via fast gradient (feat-fgrad)}.

\begin{figure}[h!]
\centering
\begin{subfigure}[t]{.325\linewidth}
\begin{center}
\includegraphics[width=0.95\linewidth]{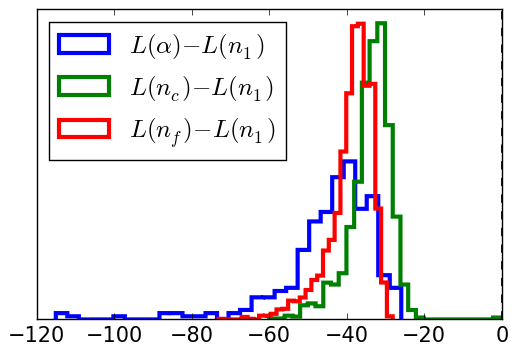}
\end{center}
\caption{\small{label-fgrad, Gaussian at $n_1(\adv)$}}
\label{fig:manifold_label}
\end{subfigure}
\begin{subfigure}[t]{.325\linewidth}
\begin{center}
\includegraphics[width=0.95\linewidth]{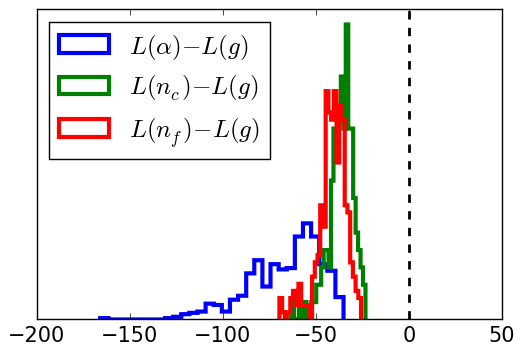}
\end{center}
\caption{\small{feat-fgrad, Gaussian at $\guide$}}
\label{fig:manifold_feat}
\end{subfigure}

\begin{subfigure}[t]{.325\linewidth}
\includegraphics[width=\linewidth]{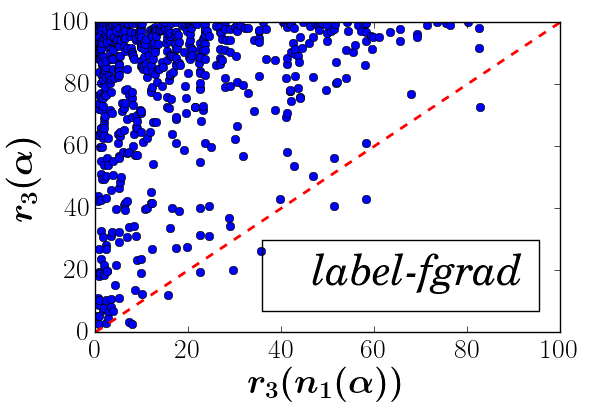}
\caption{\small{label-fgrad}}
\label{fig:rank_label}
\end{subfigure}
\begin{subfigure}[t]{.325\linewidth}
\includegraphics[width=\linewidth]{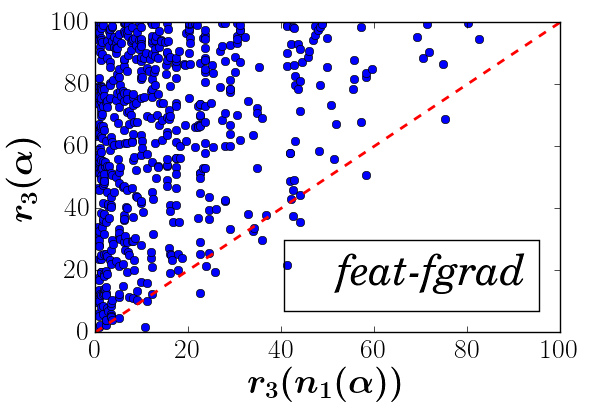}
\caption{\small{feat-fgrad}}
\label{fig:rank_feat}
\end{subfigure}

\caption{Local property analysis of label-fgrad and feat-fgrad on FC7: \ref{fig:manifold_label}-\ref{fig:manifold_feat} manifold analysis; \ref{fig:rank_label}-\ref{fig:rank_feat} neighborhood rank analysis.}
\label{fig:fgrad_plots}
\end{figure}

The same experimental setup as in Sec.~\ref{sec:comparison} is used here.  In 
Fig.~\ref{fig:fgrad_plots}, we show the nearest neighbor rank analysis and 
manifold analysis as done in Sec.~\ref{sec:natural_reps} and 
Sec.~\ref{sec:comparison}.  Moreover, 
Figs.~\ref{fig:manifold_label}-\ref{fig:manifold_feat} in compare to 
Figs.~\ref{sf:dl_fc7_train}-\ref{sf:dl_fc7_val} from {\em feature-opt} results 
and Fig.~\ref{fig:labeladv_PPCA_Dist} from {\em label-opt} results indicates 
that this adversaries are not represented as well as {\em feature-opt} by 
a Gaussian around the NN of the adversary too. Also, 
Figs.~\ref{fig:rank_label}-\ref{fig:rank_feat} in compare to 
Fig.~\ref{fig:scatter_rank} show the obvious difference in adversarial 
distribution for the same set of source and guide.

%% \begin{figure}[]
%% \centering
%% \begin{subfigure}[t]{.245\linewidth}
%% \includegraphics[width=\linewidth]{./imgs/rank_all_fgrad_label.png}
%% \caption{\small{label-fgrad}}
%% \end{subfigure}
%% \begin{subfigure}[t]{.245\linewidth}
%% \includegraphics[width=\linewidth]{./imgs/rank_all_fgrad_fc7.png}
%% \caption{\small{feat-fgrad}}
%% \end{subfigure}
%% \caption{Rank analysis for label-fgrad and feat-fgrad.}
%% \end{figure}

\subsection{Failure Cases}
There are cases in which our optimization was not successful in generating good 
adversaries.  We observed that for low resolution images or hand-drawn 
characters, the method does not always work well.  It was successful on LeNet 
with some images from MNIST or CIFAR10, but for other cases we found it 
necessary to relax the magnitude bound on the perturbations to the point that 
traces of guide images were perceptible. With Caffenet, pre-trained on ImageNet 
and then fine-tuned on the Flickr Style dataset, we could readily generate 
adversarial images using FC$8$ in the optimization (i.e., the unnormalized 
class scores), however, with FC$7$ the optimization often terminated without 
producing adversaries close to guide images.  One possible cause may be that 
the fine-tuning distorts the original natural image representation to benefit 
style classification. As a consequence, the FC$7$ layer no longer gives a good 
generic image represenation, and Euclidean distance on FC$7$ is no longer 
useful for the loss function.

% \cite{karayev2013recognizing}

\subsection{More Examples with Activation Patterns}
\begin{figure}[h!]
    \centering
\begin{subfigure}[t]{\linewidth}
{
\renewcommand{\arraystretch}{1}
\setlength\tabcolsep{2pt}

\begin{tabular}{|
>{\centering\arraybackslash}m{0.09\linewidth} |
>{\centering\arraybackslash}m{0.167\linewidth} |
>{\centering\arraybackslash}m{0.167\linewidth}
>{\centering\arraybackslash}m{0.167\linewidth}
>{\centering\arraybackslash}m{0.167\linewidth} |
>{\centering\arraybackslash}m{0.167\linewidth} | }
\hline
 & Source & $\text{FC}7$ & $\text{P}5$ & C$3$ &Guide  \\\hline  Input 
 & \includegraphics[width=\linewidth,height=.75\linewidth]{./imgs/pie.png} 
& \includegraphics[width=\linewidth,height=.75\linewidth]{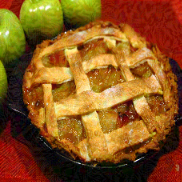} 
&
\includegraphics[width=\linewidth,height=.75\linewidth]{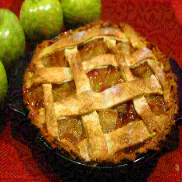} &
\includegraphics[width=\linewidth,height=.75\linewidth]{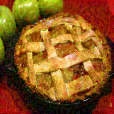} &
\includegraphics[width=\linewidth,height=.75\linewidth]{./imgs/val_43.png} \\
Inv($C3$) & 
\includegraphics[width=\linewidth,height=.75\linewidth]{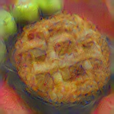} & 
\includegraphics[width=\linewidth,height=.75\linewidth]{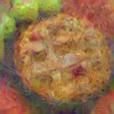} &
\includegraphics[width=\linewidth,height=.75\linewidth]{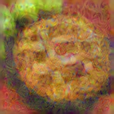} &
\includegraphics[width=\linewidth,height=.75\linewidth]{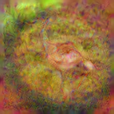} &
\includegraphics[width=\linewidth,height=.75\linewidth]{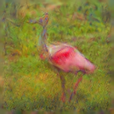} 
\\
Inv($P5$) & 
\includegraphics[width=\linewidth,height=.75\linewidth]{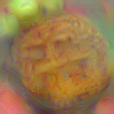} & 
\includegraphics[width=\linewidth,height=.75\linewidth]{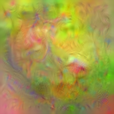} &
\includegraphics[width=\linewidth,height=.75\linewidth]{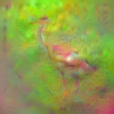} &
\includegraphics[width=\linewidth,height=.75\linewidth]{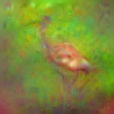} &
\includegraphics[width=\linewidth,height=.75\linewidth]{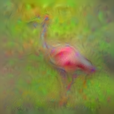} 
\\
Inv($FC7$) & 
\includegraphics[width=\linewidth,height=.75\linewidth]{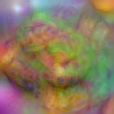} & 
\includegraphics[width=\linewidth,height=.75\linewidth]{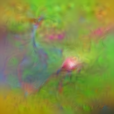} &
\includegraphics[width=\linewidth,height=.75\linewidth]{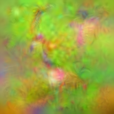} &
\includegraphics[width=\linewidth,height=.75\linewidth]{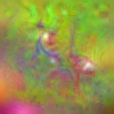} &
\includegraphics[width=\linewidth,height=.75\linewidth]{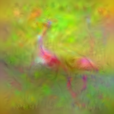} 
\\

\hline
\end{tabular}
}
\end{subfigure}
\vspace*{0.2cm}

\begin{subfigure}[t]{\linewidth}
{
\centering
\renewcommand{\arraystretch}{1}
\setlength\tabcolsep{.1pt}
\begin{tabular}{
|>{\centering\arraybackslash}m{0.205\linewidth}
>{\centering\arraybackslash}m{0.205\linewidth}
>{\centering\arraybackslash}m{0.205\linewidth}|
>{\centering\arraybackslash}m{0.125\linewidth}
>{\centering\arraybackslash}m{0.125\linewidth}
>{\centering\arraybackslash}m{0.125\linewidth}|
}
\hline
\includegraphics[width=\linewidth]{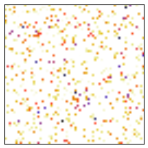} &
\includegraphics[width=\linewidth]{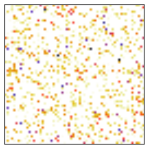} &
\includegraphics[width=\linewidth]{./imgs/f7_43.png} &
\includegraphics[height=\linewidth, angle=90]{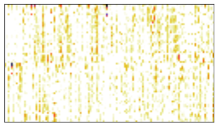} &
\includegraphics[height=\linewidth, angle=90]{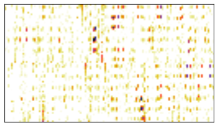} &
\includegraphics[height=\linewidth, angle=90]{./imgs/p5_43.png}\\
Source & FC7 Advers. & Guide & Source & P5 Advers. & Guide \\ \hline
%\multicolumn{3}{|c|}{FC$7$ Activations} & \multicolumn{3}{c|}{ P$5$ 
%Activations  } \\
%\hline
\end{tabular}
}
\end{subfigure}
\caption{
    Inverted images and activation plot for a pair of source and guide image 
    shown in the first row (Input). This figure has same setting as 
    Fig.~\ref{fig:adv_invert}.
}

\label{fig:adv_invert2}
\end{figure}

\begin{figure*}[h!]

\begin{subfigure}[t]{\linewidth}
{
\renewcommand{\arraystretch}{1}
\setlength\tabcolsep{2pt}
\begin{tabular}{|
>{\centering\arraybackslash}m{0.09\linewidth} |
>{\centering\arraybackslash}m{0.167\linewidth} |
>{\centering\arraybackslash}m{0.167\linewidth}
>{\centering\arraybackslash}m{0.167\linewidth}
>{\centering\arraybackslash}m{0.167\linewidth} |
>{\centering\arraybackslash}m{0.167\linewidth} | }
\hline
& Source & $\text{FC}7$ & $\text{P}5$ & C$3$ &Guide  \\\hline  
Input & 
\includegraphics[width=\linewidth,height=.75\linewidth]{./imgs/pie.png} & 
\includegraphics[width=\linewidth,height=.75\linewidth]{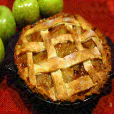} &
\includegraphics[width=\linewidth,height=.75\linewidth]{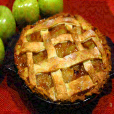} &
\includegraphics[width=\linewidth,height=.75\linewidth]{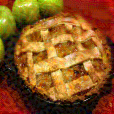} &
\includegraphics[width=\linewidth,height=.75\linewidth]{./imgs/27.png} \\
Inv(C$3$) & 
\includegraphics[width=\linewidth,height=.75\linewidth]{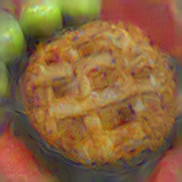} & 
\includegraphics[width=\linewidth,height=.75\linewidth]{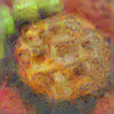} &
\includegraphics[width=\linewidth,height=.75\linewidth]{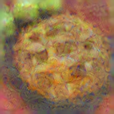} &
\includegraphics[width=\linewidth,height=.75\linewidth]{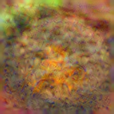} &
\includegraphics[width=\linewidth,height=.75\linewidth]{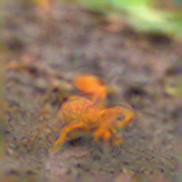} 
\\
Inv($\text{P}5$) & 
\includegraphics[width=\linewidth,height=.75\linewidth]{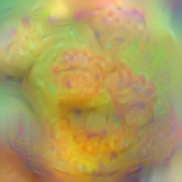} & 
\includegraphics[width=\linewidth,height=.75\linewidth]{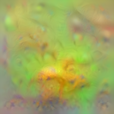} &
\includegraphics[width=\linewidth,height=.75\linewidth]{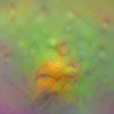} &
\includegraphics[width=\linewidth,height=.75\linewidth]{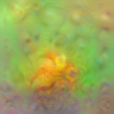} &
\includegraphics[width=\linewidth,height=.75\linewidth]{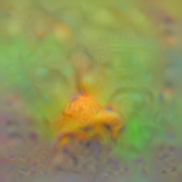} 
\\
Inv($\text{FC}7$) & 
\includegraphics[width=\linewidth,height=.75\linewidth]{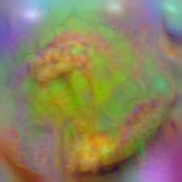} & 
\includegraphics[width=\linewidth,height=.75\linewidth]{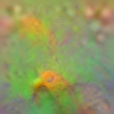} &
\includegraphics[width=\linewidth,height=.75\linewidth]{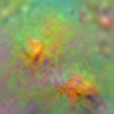} &
\includegraphics[width=\linewidth,height=.75\linewidth]{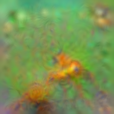} &
\includegraphics[width=\linewidth,height=.75\linewidth]{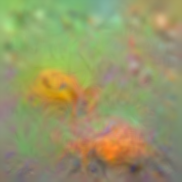} 
\\
\hline
\end{tabular}
}
\end{subfigure}
\vspace*{0.2cm}

\begin{subfigure}[t]{\linewidth}
{
\centering
\renewcommand{\arraystretch}{1}
\setlength\tabcolsep{.1pt}
\begin{tabular}{
|>{\centering\arraybackslash}m{0.205\linewidth}
>{\centering\arraybackslash}m{0.205\linewidth}
>{\centering\arraybackslash}m{0.205\linewidth}|
>{\centering\arraybackslash}m{0.125\linewidth}
>{\centering\arraybackslash}m{0.125\linewidth}
>{\centering\arraybackslash}m{0.125\linewidth}|
}
\hline
\includegraphics[width=\linewidth]{./imgs/f7_pie.png} &
\includegraphics[width=\linewidth]{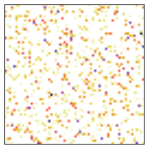} &
\includegraphics[width=\linewidth]{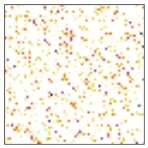} &
\includegraphics[height=\linewidth, angle=90]{./imgs/p5_pie.png} &
\includegraphics[height=\linewidth, angle=90]{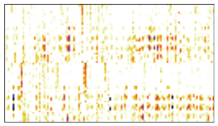} &
\includegraphics[height=\linewidth, angle=90]{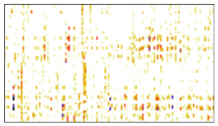}\\
Source & FC7 Advers. & Guide & Source & P5 Advers. & Guide \\ \hline
%\multicolumn{3}{|c|}{FC$7$ Activations} & \multicolumn{3}{c|}{ P$5$ 
%Activations  } \\
%\hline
\end{tabular}
}
\end{subfigure}
\caption{
    Inverted images and activation plot for a pair of source and guide image 
    shown in the first row (Input). This figure has same setting as 
    Fig.~\ref{fig:adv_invert}.
}
\label{fig:adv_invert4}
\end{figure*}

\begin{figure}[h!]
    \centering
\begin{subfigure}[t]{\linewidth}
{
\renewcommand{\arraystretch}{1}
\setlength\tabcolsep{2pt}
\begin{tabular}{|
>{\centering\arraybackslash}m{0.09\linewidth} |
>{\centering\arraybackslash}m{0.167\linewidth} |
>{\centering\arraybackslash}m{0.167\linewidth}
>{\centering\arraybackslash}m{0.167\linewidth}
>{\centering\arraybackslash}m{0.167\linewidth} |
>{\centering\arraybackslash}m{0.167\linewidth} | }
\hline
& Source & $\text{FC}7$ & $\text{P}5$ & C$3$ &Guide  \\\hline Input 
& \includegraphics[width=\linewidth,height=.75\linewidth]{./imgs/train_730.png} 
& \includegraphics[width=\linewidth,height=.75\linewidth]{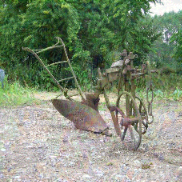} 
&
\includegraphics[width=\linewidth,height=.75\linewidth]{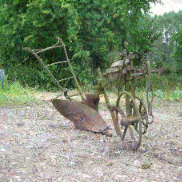} &
\includegraphics[width=\linewidth,height=.75\linewidth]{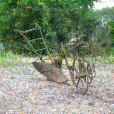} &
\includegraphics[width=\linewidth,height=.75\linewidth]{./imgs/27.png} \\
Inv($C3$) 
& \includegraphics[width=\linewidth,height=.75\linewidth]{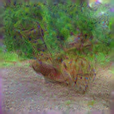} 
& \includegraphics[width=\linewidth,height=.75\linewidth]{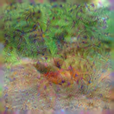} 
&
\includegraphics[width=\linewidth,height=.75\linewidth]{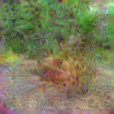} &
\includegraphics[width=\linewidth,height=.75\linewidth]{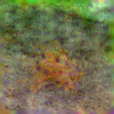} &
\includegraphics[width=\linewidth,height=.75\linewidth]{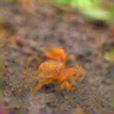} 
\\
Inv($P5$) 
& \includegraphics[width=\linewidth,height=.75\linewidth]{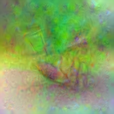} 
& \includegraphics[width=\linewidth,height=.75\linewidth]{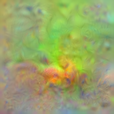} 
&
\includegraphics[width=\linewidth,height=.75\linewidth]{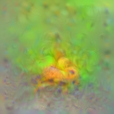} &
\includegraphics[width=\linewidth,height=.75\linewidth]{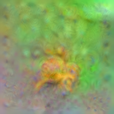} &
\includegraphics[width=\linewidth,height=.75\linewidth]{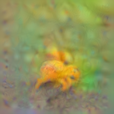} 
\\
Inv($FC7$)
& \includegraphics[width=\linewidth,height=.75\linewidth]{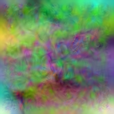} 
& \includegraphics[width=\linewidth,height=.75\linewidth]{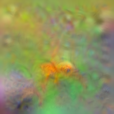} 
&
\includegraphics[width=\linewidth,height=.75\linewidth]{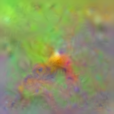} &
\includegraphics[width=\linewidth,height=.75\linewidth]{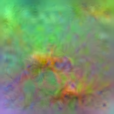} &
\includegraphics[width=\linewidth,height=.75\linewidth]{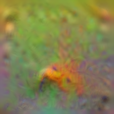} 
\\
\hline
\end{tabular}
}
\end{subfigure}

\vspace*{0.2cm}

\begin{subfigure}[t]{\linewidth}
{
\centering
\renewcommand{\arraystretch}{1}
\setlength\tabcolsep{.1pt}
\begin{tabular}{
|>{\centering\arraybackslash}m{0.205\linewidth}
>{\centering\arraybackslash}m{0.205\linewidth}
>{\centering\arraybackslash}m{0.205\linewidth}|
>{\centering\arraybackslash}m{0.125\linewidth}
>{\centering\arraybackslash}m{0.125\linewidth}
>{\centering\arraybackslash}m{0.125\linewidth}|
}
\hline
\includegraphics[width=\linewidth]{./imgs/f7_730.png} &
\includegraphics[width=\linewidth]{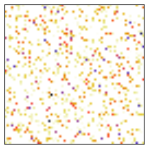} &
\includegraphics[width=\linewidth]{./imgs/f7_27.png} &
\includegraphics[height=\linewidth, angle=90]{./imgs/p5_730.png} &
\includegraphics[height=\linewidth, angle=90]{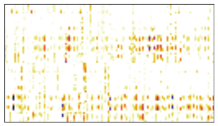} &
\includegraphics[height=\linewidth, angle=90]{./imgs/p5_27.png}\\
Source & FC7 Advers. & Guide & Source & P5 Advers. & Guide \\ \hline
%\multicolumn{3}{|c|}{FC$7$ Activations} & \multicolumn{3}{c|}{ P$5$ 
%Activations  } \\
%\hline
\end{tabular}
}
\end{subfigure}
\caption{
    Inverted images and activation plot for a pair of source and guide image 
    shown in the first row (Input). This figure has same setting as 
    Fig.~\ref{fig:adv_invert}.
}
\label{fig:adv_invert3}
\end{figure}

\begin{figure}[h]
    \centering
\begin{subfigure}[t]{\linewidth}
{
\renewcommand{\arraystretch}{1}
\setlength\tabcolsep{2pt}

\begin{tabular}{|
>{\centering\arraybackslash}m{0.09\linewidth} |
>{\centering\arraybackslash}m{0.167\linewidth} |
>{\centering\arraybackslash}m{0.167\linewidth}
>{\centering\arraybackslash}m{0.167\linewidth}
>{\centering\arraybackslash}m{0.167\linewidth} |
>{\centering\arraybackslash}m{0.167\linewidth} | }
\hline
 & Source & $\text{FC}7$ & $\text{P}5$ & C$3$ &Guide  \\\hline
 Input 
 & \includegraphics[width=\linewidth,height=.75\linewidth]{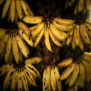} 
 & \includegraphics[width=\linewidth,height=.75\linewidth]{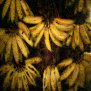} 
 &
\includegraphics[width=\linewidth,height=.75\linewidth]{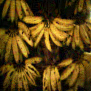} &
\includegraphics[width=\linewidth,height=.75\linewidth]{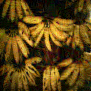} &
\includegraphics[width=\linewidth,height=.75\linewidth]{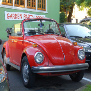} \\
Inv($C3$) & 
\includegraphics[width=\linewidth,height=.75\linewidth]{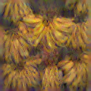} & 
\includegraphics[width=\linewidth,height=.75\linewidth]{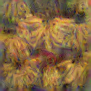} &
\includegraphics[width=\linewidth,height=.75\linewidth]{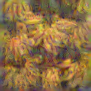} &
\includegraphics[width=\linewidth,height=.75\linewidth]{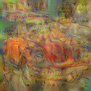} &
\includegraphics[width=\linewidth,height=.75\linewidth]{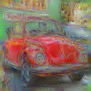} 
\\
Inv($P5$) & 
\includegraphics[width=\linewidth,height=.75\linewidth]{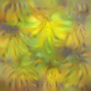} & 
\includegraphics[width=\linewidth,height=.75\linewidth]{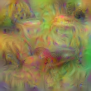} &
\includegraphics[width=\linewidth,height=.75\linewidth]{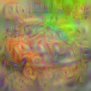} &
\includegraphics[width=\linewidth,height=.75\linewidth]{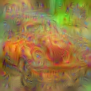} &
\includegraphics[width=\linewidth,height=.75\linewidth]{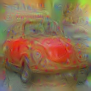} 
\\
Inv($FC7$) & 
\includegraphics[width=\linewidth,height=.75\linewidth]{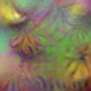} & 
\includegraphics[width=\linewidth,height=.75\linewidth]{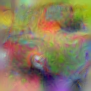} &
\includegraphics[width=\linewidth,height=.75\linewidth]{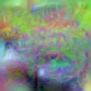} &
\includegraphics[width=\linewidth,height=.75\linewidth]{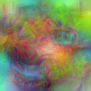} &
\includegraphics[width=\linewidth,height=.75\linewidth]{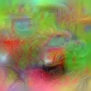} 
\\
 
 \hline
\end{tabular}
}
\end{subfigure}
\vspace*{0.2cm}

\begin{subfigure}[t]{\linewidth}
{
\centering
\renewcommand{\arraystretch}{1}
\setlength\tabcolsep{.1pt}
\begin{tabular}{
|>{\centering\arraybackslash}m{0.205\linewidth}
>{\centering\arraybackslash}m{0.205\linewidth}
>{\centering\arraybackslash}m{0.205\linewidth}|
>{\centering\arraybackslash}m{0.125\linewidth}
>{\centering\arraybackslash}m{0.125\linewidth}
>{\centering\arraybackslash}m{0.125\linewidth}|
}
\hline
\includegraphics[width=\linewidth]{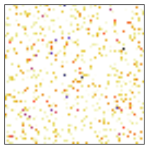} &
\includegraphics[width=\linewidth]{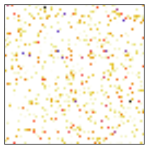} &
\includegraphics[width=\linewidth]{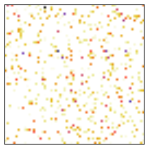} &
\includegraphics[height=\linewidth, angle=90]{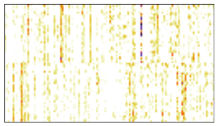} &
\includegraphics[height=\linewidth, angle=90]{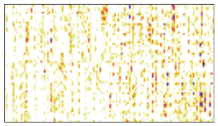} &
\includegraphics[height=\linewidth, angle=90]{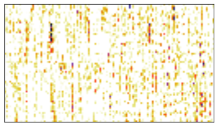}\\
Source & FC7 Advers. & Guide & Source & P5 Advers. & Guide \\ \hline
%\multicolumn{3}{|c|}{FC$7$ Activations} & \multicolumn{3}{c|}{ P$5$ 
%Activations  } \\
%\hline
\end{tabular}
}
\end{subfigure}
\caption{
    Inverted images and activation plot for a pair of source and guide image 
    shown in the first row (Input). This figure has same setting as 
    Fig.~\ref{fig:adv_invert}.
}
\label{fig:adv_invert5}
\end{figure}

\begin{figure}[h!]
    \centering

\begin{subfigure}[t]{\linewidth}
{
\renewcommand{\arraystretch}{1}
\setlength\tabcolsep{2pt}
\begin{tabular}{|
>{\centering\arraybackslash}m{0.09\linewidth} |
>{\centering\arraybackslash}m{0.167\linewidth} |
>{\centering\arraybackslash}m{0.167\linewidth}
>{\centering\arraybackslash}m{0.167\linewidth}
>{\centering\arraybackslash}m{0.167\linewidth} |
>{\centering\arraybackslash}m{0.167\linewidth} | }
\hline
& Source & $\text{FC}7$ & $\text{P}5$ & C$3$ &Guide  \\\hline
Input 
& \includegraphics[width=\linewidth,height=.75\linewidth]{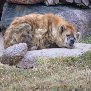} 
& \includegraphics[width=\linewidth,height=.75\linewidth]{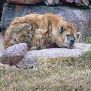} 
&
\includegraphics[width=\linewidth,height=.75\linewidth]{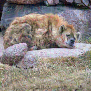} &
\includegraphics[width=\linewidth,height=.75\linewidth]{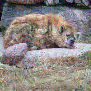} &
\includegraphics[width=\linewidth,height=.75\linewidth]{./imgs/car.png} \\
Inv($C3$) & 
\includegraphics[width=\linewidth,height=.75\linewidth]{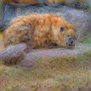} & 
\includegraphics[width=\linewidth,height=.75\linewidth]{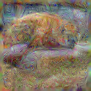} &
\includegraphics[width=\linewidth,height=.75\linewidth]{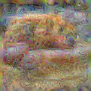} &
\includegraphics[width=\linewidth,height=.75\linewidth]{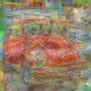} &
\includegraphics[width=\linewidth,height=.75\linewidth]{./imgs/car/l10-recon.png} 
\\
Inv($P5$) & 
\includegraphics[width=\linewidth,height=.75\linewidth]{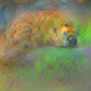} & 
\includegraphics[width=\linewidth,height=.75\linewidth]{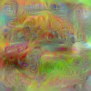} &
\includegraphics[width=\linewidth,height=.75\linewidth]{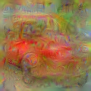} &
\includegraphics[width=\linewidth,height=.75\linewidth]{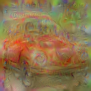} &
\includegraphics[width=\linewidth,height=.75\linewidth]{./imgs/car/l15-recon.png} 
\\
Inv($FC7$) & 
\includegraphics[width=\linewidth,height=.75\linewidth]{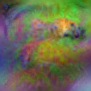} & 
\includegraphics[width=\linewidth,height=.75\linewidth]{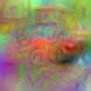} &
\includegraphics[width=\linewidth,height=.75\linewidth]{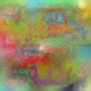} &
\includegraphics[width=\linewidth,height=.75\linewidth]{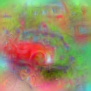} &
\includegraphics[width=\linewidth,height=.75\linewidth]{./imgs/car/l19-recon.png} 
\\

\hline
\end{tabular}
}
\end{subfigure}
\vspace*{0.2cm}

\begin{subfigure}[t]{\linewidth}
{
\centering
\renewcommand{\arraystretch}{1}
\setlength\tabcolsep{.1pt}
\begin{tabular}{
|>{\centering\arraybackslash}m{0.205\linewidth}
>{\centering\arraybackslash}m{0.205\linewidth}
>{\centering\arraybackslash}m{0.205\linewidth}|
>{\centering\arraybackslash}m{0.125\linewidth}
>{\centering\arraybackslash}m{0.125\linewidth}
>{\centering\arraybackslash}m{0.125\linewidth}|
}
\hline
\includegraphics[width=\linewidth]{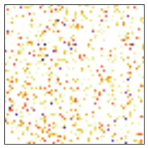} &
\includegraphics[width=\linewidth]{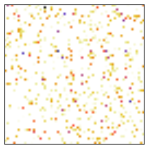} &
\includegraphics[width=\linewidth]{./imgs/f7_car.png} &
\includegraphics[height=\linewidth, angle=90]{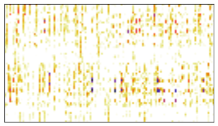} &
\includegraphics[height=\linewidth, angle=90]{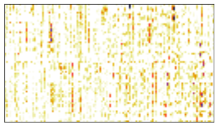} &
\includegraphics[height=\linewidth, angle=90]{./imgs/p5_car.png}\\
Source & FC7 Advers. & Guide & Source & P5 Advers. & Guide \\ \hline
%\multicolumn{3}{|c|}{FC$7$ Activations} & \multicolumn{3}{c|}{ P$5$ 
%Activations  } \\
%\hline
\end{tabular}
}
\end{subfigure}
\caption{
    Inverted images and activation plot for a pair of source and guide image 
    shown in the first row (Input). This figure has same setting as 
    Fig.~\ref{fig:adv_invert}.
}
\label{fig:adv_invert6}
\end{figure}

Finally, we dedicate the remaining pages to several pairs of source and guide
along with their adversaries, activation patterns and inverted images as a 
complementary to Fig.~\ref{fig:adv_invert}. Figs.~\ref{fig:adv_invert2}, 
\ref{fig:adv_invert4}, \ref{fig:adv_invert3}, \ref{fig:adv_invert5} and \ref{fig:adv_invert6}
all have similar setup as it is discussed in Sec.~\ref{method}.

%\begin{subfigure}[t]{\linewidth}
%{
%\centering
%\renewcommand{\arraystretch}{1}
%\setlength\tabcolsep{.1pt}
%\begin{tabular}{
%|>{\centering\arraybackslash}m{0.205\linewidth}
%>{\centering\arraybackslash}m{0.205\linewidth}
%>{\centering\arraybackslash}m{0.205\linewidth}|
%>{\centering\arraybackslash}m{0.125\linewidth}
%>{\centering\arraybackslash}m{0.125\linewidth}
%>{\centering\arraybackslash}m{0.125\linewidth}|
%}
%\hline
%\includegraphics[width=\linewidth]{./imgs/f7_730.png} &
%\includegraphics[width=\linewidth]{./imgs/f7_730_27.png} &
%\includegraphics[width=\linewidth]{./imgs/f7_27.png} &
%\includegraphics[height=\linewidth, angle=90]{./imgs/p5_730.png} &
%\includegraphics[height=\linewidth, angle=90]{./imgs/p5_730_27.png} &
%\includegraphics[height=\linewidth, angle=90]{./imgs/p5_27.png}\\
%Source & FC7 Advers. & Guide & Source & P5 Advers. & Guide \\ \hline
%%\multicolumn{3}{|c|}{FC$7$ Activations} & \multicolumn{3}{c|}{ P$5$ 
%%Activations  } \\
%%\hline
%\end{tabular}
%}
%\end{subfigure}
%\caption{
%    Inverted images and activation plot for a pair of source and guide image 
%    shown in the first row (Input). This figure has same setting as 
%    Fig.~\ref{fig:adv_invert}.
%}
%\label{fig:adv_invert5}
%\end{figure}
%
%

\end{document}